%% file: main.tex
\documentclass{article}
\usepackage{xcolor}         
\usepackage[preprint]{style}

\usepackage[utf8]{inputenc} 
\usepackage[T1]{fontenc}    
\usepackage{hyperref}       
\usepackage{url}            
\usepackage{booktabs}       
\usepackage{amsfonts}       
\usepackage{nicefrac}       
\usepackage{microtype}      
\usepackage{enumitem}
\usepackage{times}
\usepackage{latexsym}
\usepackage{graphicx}
\usepackage{multicol}
\usepackage{multirow}
\usepackage{amsmath}
\usepackage{amssymb}
\usepackage{bbding}
\usepackage{subfigure}
\usepackage{colortbl}
\usepackage{pifont}
\usepackage{makecell}
\usepackage{mathrsfs}
\usepackage{bm}
\usepackage{bbm}
\usepackage{dsfont}
\usepackage{pgf}
\usepackage{fontawesome}
\usepackage{wrapfig}
\usepackage{tcolorbox}
\usepackage{array}
\usepackage{longtable} 
\usepackage{caption}

\definecolor{darkgray}{gray}{0.3}
\definecolor{lightblue}{RGB}{115, 192, 222}

\newtcolorbox{promptbox}[2][Prompt]{
colback=black!5!white,
arc=4pt, 
boxrule=0.5pt,
fonttitle=\bfseries,
title=#1, 
before upper={\small}, fontupper=\fontfamily{ptm}\selectfont,
colframe=#2,
}

\newcommand{\ourmethod}{xDebugGen}
\newcommand{\baseline}{xDebugCoder}
\newcommand{\benchmark}{\textsc{MdEval}}
\newcommand{\instruct}{\textsc{MdEval-Instruct}}

\newcommand{\bmap}{{\raisebox{-0.1em}{\includegraphics[height=0.9em]{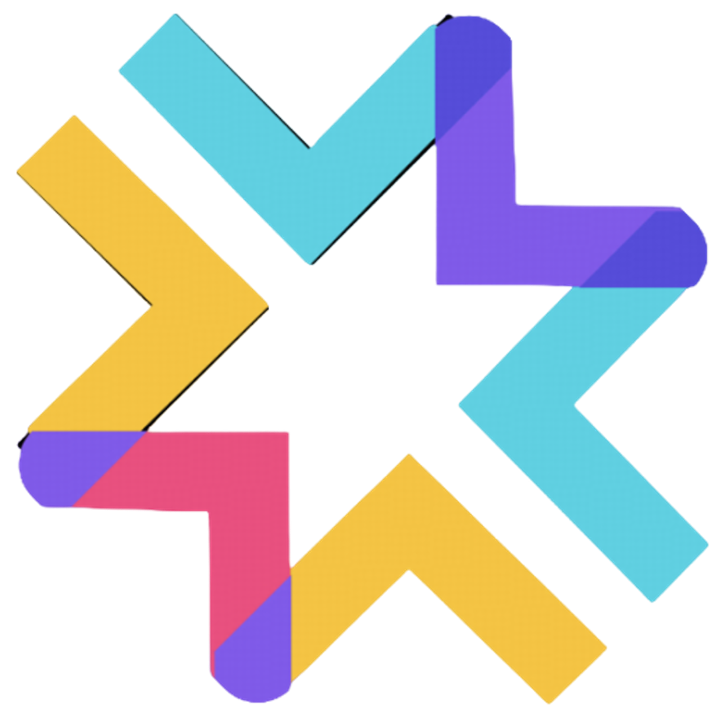}}}}

\title{\benchmark{}: Massively Multilingual Code Debugging}

\author{
  Shukai Liu\textsuperscript{\rm 1 \thanks{\ \ Equal contribution. }}, 
  Linzheng Chai\textsuperscript{\rm 1 *},
  Jian Yang\textsuperscript{\rm 1 *}\thanks{\ \ Corresponding Author.},  
  {\bf Jiajun Shi}\textsuperscript{\rm 1}, 
  {\bf He Zhu}\textsuperscript{\rm 1}, 
  {\bf Liran Wang}\textsuperscript{\rm 1}, 
  {\bf Ke Jin}\textsuperscript{\rm 1}, \\
  {\bf Wei Zhang}\textsuperscript{$^\bmap$},  
  {\bf Hualei Zhu}\textsuperscript{\rm 1},
  {\bf Shuyue Guo}\textsuperscript{$^\bmap$},
  {\bf Tao Sun}\textsuperscript{\rm 1}, 
  {\bf Jiaheng Liu}\textsuperscript{\rm 1}, 
  {\bf Yunlong Duan}\textsuperscript{$^\bmap$}, \\
  {\bf Yu Hao}\textsuperscript{$^\bmap$},
  {\bf Liqun Yang}\textsuperscript{\rm 1}, 
  {\bf Guanglin Niu}\textsuperscript{\rm 1}, 
  {\bf Ge Zhang}\textsuperscript{$^\bmap$},
  {\bf Zhoujun Li}\textsuperscript{\rm 1} \\
 \textsuperscript{\rm 1}CCSE, Beihang University,~\textsuperscript{\rm 2}$^\bmap${\rm{M-A-P}} \\
  \texttt{liusk,chailinzheng@buaa.edu.cn} \\
}

\begin{document}

\maketitle

\begin{abstract}
Code large language models (LLMs) have made significant progress in code debugging by directly generating the correct code based on the buggy code snippet. Programming benchmarks, typically consisting of buggy code snippets and their associated test cases, are used to assess the debugging capabilities of LLMs. However, many existing benchmarks primarily focus on Python and are often limited in terms of language diversity (e.g., DebugBench and DebugEval). To advance the field of multilingual debugging with LLMs, we propose the first massively multilingual debugging benchmark, which includes 3.9K test samples of 20 programming languages and covers the automated program repair (APR) task, the bug localization(BL) task, and the bug identification (BI) task. In addition, we introduce the debugging instruction corpora \instruct{} by injecting bugs into the correct multilingual queries and solutions (\ourmethod{}).
Further, a multilingual debugger \baseline{} trained on \instruct{} as a strong baseline specifically to handle bugs of a wide range of programming languages (e.g. ``Missing Mut'' in language Rust and ``Misused Macro Definition'' in language C). Our extensive experiments on \benchmark{} reveal a notable performance gap between open-source models and closed-source LLMs (e.g., GPT and Claude series), highlighting huge room for improvement in multilingual code debugging scenarios\footnote{\url{https://mdeval-code.github.io/}}.
\end{abstract}

\section{Introduction}
\label{sec:introduction}
Large language models (LLMs)~\citep{gpt4, llama,qwen2} designed for code, such as CodeLlama \citep{code_llama}, DeepSeekCoder \citep{deepseek_coder}, and QwenCoder \citep{qwencoder}, are highly effective in code understanding and generation. These capabilities make them particularly useful for debugging, where deep comprehension of code structure and logic is essential. Automated program repair (APR) \citep{wen2024fixing} aims to automatically fix bugs without human involvement, significantly reducing time and costs in development processes.

\begin{figure}[t]
\centering
\includegraphics[width=1\linewidth]{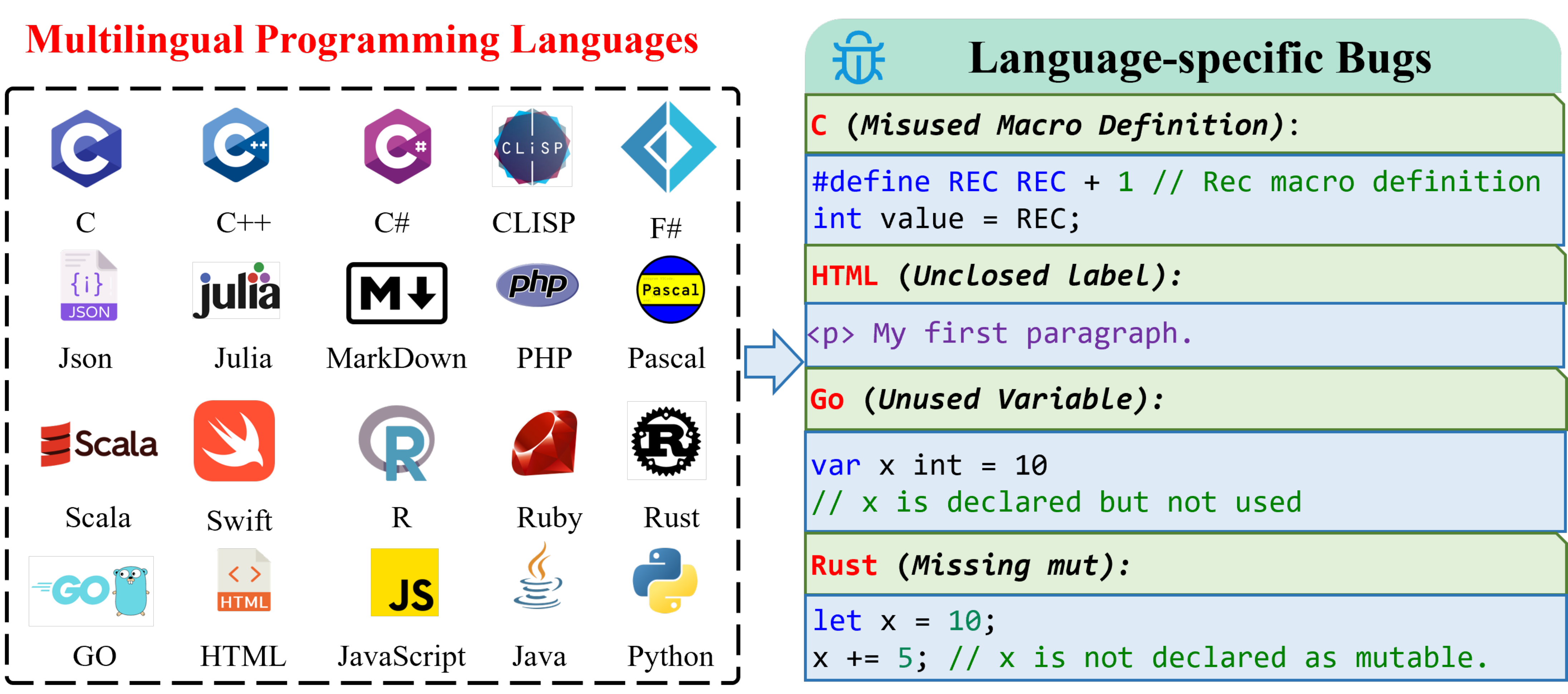}
\caption{Massively multilingual evaluation task comprised of three tasks, including code generation, code completion, and code explanation.}
\vspace{-15pt}
\label{fig:intro}
\end{figure}

LLMs have recently shown considerable potential in this area. For instance, CodeX~\citep{codex} and GPT-4 series~\citep{gpt4} outperforming previous conventional methods have demonstrated promising results on bug benchmarks such as QuixBugs~\citep{quixbugs}. The recent work DebugBench~\citep{debugbench} creates a debugging benchmark including Python, Java, and CPP for LLM evaluation. However, for the diverse programming languages in Figure \ref{fig:intro}, the multilingual debugging scenario poses more language-specific challenges for APR. Multilingual issues (e.g. ``Misused Macro Definition'' in programming language C, ``Missing mut'' in Rust, and ``Unused Variable'' in Go) highlight the complexities and diversities of locating and fixing bugs in the multilingual debugging scenario. Therefore, there is an urgent need to build a truly massively multilingual debugging code benchmark with a wide variety of generic and language-specific bug types. 

To further characterize the debugging performance of LLMs across different programming languages, we introduce \benchmark{}, a framework for data construction, evaluation benchmark, and a multilingual debugging baseline \baseline{}, to advance the development of code debugging.
First, we propose \benchmark{}, the first massively multilingual evaluation benchmark for code debugging covering 20 programming languages and 3.9K samples to assess the capabilities of LLMs across a wide range of languages. Further, we create \instruct{}, a multilingual debugging instruction corpus in 20 languages to help the LLM fix the bug given the buggy code snippet. Besides, we propose \ourmethod{} to create the buggy and correct code pair for debugging instruction tuning. The bugs are injected into the queries and solutions with our designed three strategies (1) Injecting bugs into query. (2) Injecting bugs into solution. (3) Injecting bugs with the round-trip code translation. Leveraging \instruct{}, we develop \baseline{} as a strong baseline, assessing the transferability of LLMs in multilingual debugging tasks.

The contributions are summarized as follows: (1) We propose \benchmark{}, a comprehensive multilingual code debugging benchmark consisting of 3.9K samples spanning three tasks: automated program repair (APR), code localization (BL), and bug identification (BI). This benchmark covers 20 languages and includes both generic and language-specific bug types.
(2) We introduce the massively multilingual code debugging instruction corpora \instruct{} created by \ourmethod{}. By injecting bugs into the correct multilingual query or response, we can create pairs of buggy code and the correct code for instruction tuning.
(3) We systematically evaluate the multilingual code debugging capabilities of 40 models on our created \benchmark{} and create a leaderboard to evaluate them on 20 programming languages dynamically. Notably, extensive experiments suggest that comprehensive multilingual multitask evaluation can realistically measure the gap between open-source (e.g. DeepSeekCoder and Qwen-Coder) and closed-source models (e.g. Claude series).

\section{Multilingual Code Debugging Evaluation: \benchmark{}}
\subsection{Data Overview}

In \autoref{tab:detail_data}, the \benchmark{} consists of 3.9K problems. Following~\citet{debugeval}, we design 3 multilingual debugging-related tasks: Automated Program Repair, Bug Localization, and Bug Identification. Each task contains about 1.3K questions, with more than 60 problems in each language. 
Each problem in \benchmark{} includes \textit{question, example test cases, buggy code, correct code, and unit tests}.

\begin{wraptable}[14]{r}{5.0cm}

\scalebox{0.65}{
\begin{tabular}{lr}
\hline
\toprule
\textbf{Statistics}            & \textbf{Number}         \\
\midrule
\textbf{Problems}              & $3,897$                 \\
Automated Program Repair       & $1,299$                 \\
Bug Localization               & $1,299$                 \\
Bug Identification             & $1,299$                 \\
Total Test Cases               & $7,133$                 \\
\midrule
\textbf{\#Difficulty Level } &                          \\
- Easy/Medium/Hard               & $1,146$/$1,407$/$1362$ \\
\midrule
\textbf{Length}                                         \\
Question \\
~~~~- \textit{maximum length}   & $291$ tokens          \\
~~~~- \textit{minimum length}   & $7$ tokens            \\
~~~~- \textit{avg length}       & $70$ tokens         \\
Buggy code  \\
~~~~- \textit{maximum length}   & $19,265$ tokens          \\
~~~~- \textit{minimum length}   & $15$ tokens            \\
~~~~- \textit{avg length}       & $320.6$ tokens         \\
\bottomrule
\hline
\end{tabular}}
\caption{\benchmark{} dataset statistics. }
\label{tab:detail_data}
\end{wraptable}
We calculate the length of the question and buggy code using the CodeLlama tokenizer \citep{code_llama}. The average question length is 83 words, highlighting their detailed descriptive nature. The average buggy code length is 239 tokens, indicating the complexity of the code. In addition, the total number of unit tests for the dataset is 6,838, to ensure the accuracy of the bug-fix judgment.

\begin{figure*}[htb]
	\centering
	\subfigure[Generic Error Types]{\includegraphics[width=0.4\textwidth]{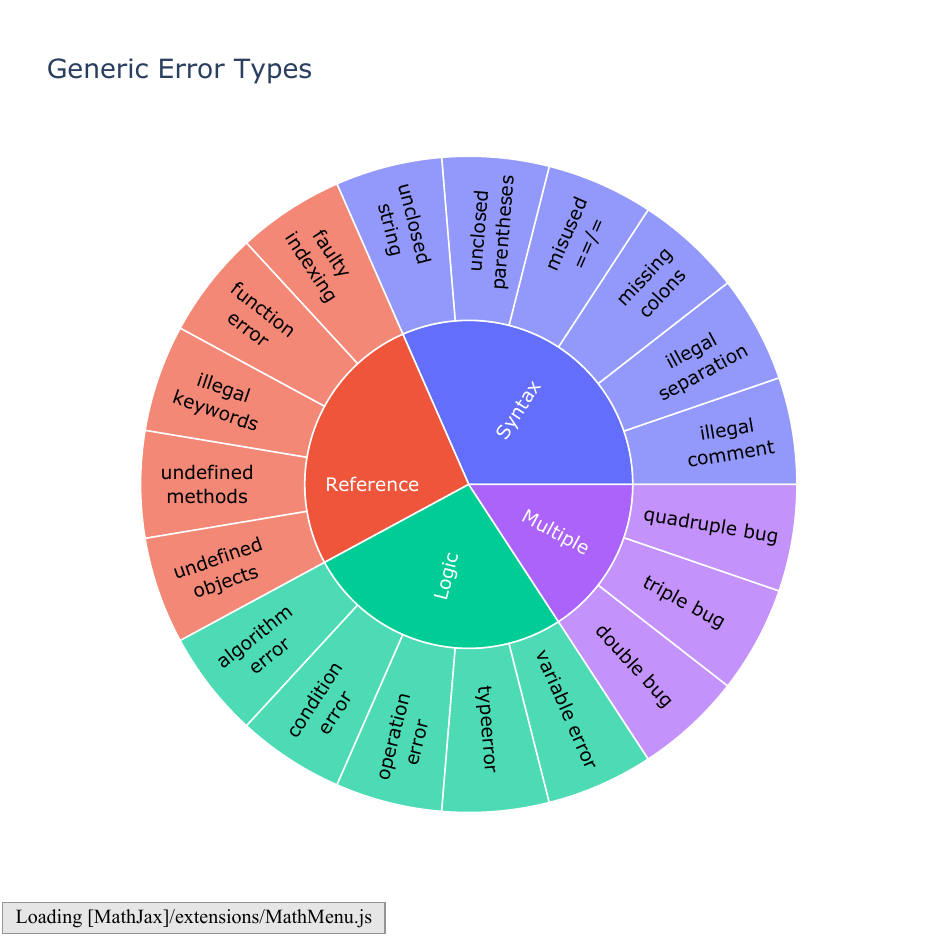}}
    \hspace{0.1\textwidth} 
	\subfigure[Language-Specific Error Types]{\includegraphics[width=0.4\textwidth]{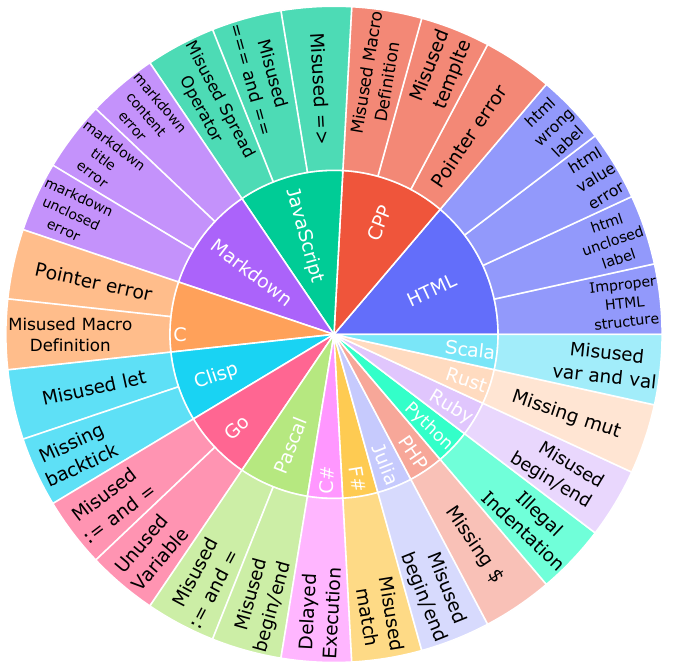}}
	\caption{Error types in \benchmark{}. Part~(a) shows generic error types, and Part~(b) lists language-specific error types.}
	\label{fig:error_types}
\end{figure*}

\begin{figure*}[h]
\begin{center}
    \includegraphics[width=1.0\textwidth]{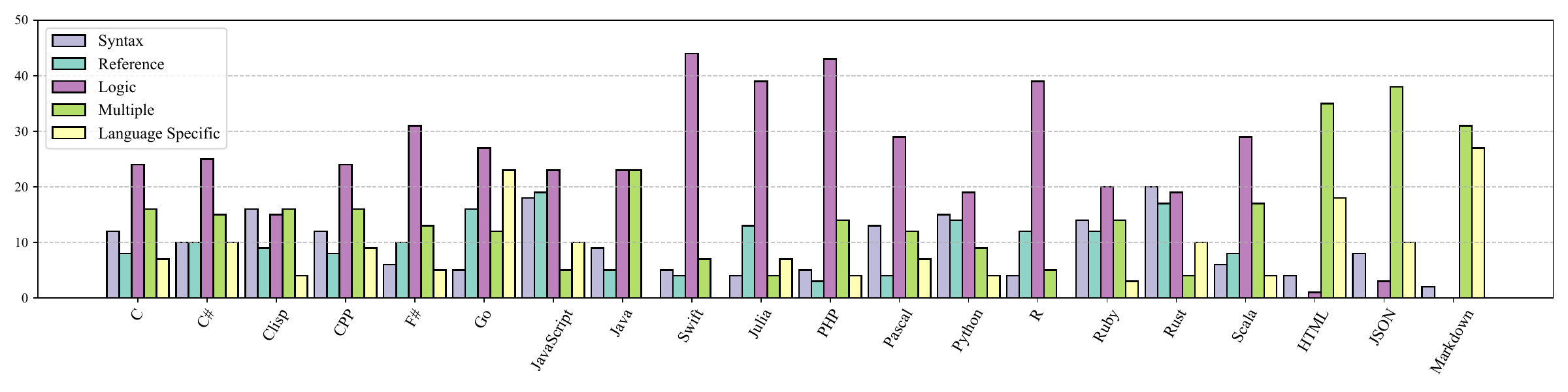}
    \vspace{-5pt}
    \caption{Error types distribution in 20 programming Languages from the \benchmark{}. }
    \label{fig:data_statistics}
    \vspace{-10pt}
\end{center}
\end{figure*}

In \autoref{tab:compare_bench}, we compare \benchmark{} with other code debugging benchmarks. Our benchmark provides a valuable enhancement to existing ones, significantly expanding the variety of programming languages and introducing language-specific error types, along with a greater number of questions and diverse bug-fixing tasks. The error types in \benchmark{} are shown in \autoref{fig:error_types}. \autoref{fig:data_statistics} plots error types distribution. We strive to cover all error types in each language. Due to the inherent differences among languages, we ensure a balanced distribution of difficulty levels, leading to variations in the distribution of error types across languages.
\begin{table*}[h]
    \centering
    \caption{Comparison between \benchmark{} and other code debugging benchmarks. \benchmark{} significantly provides a comprehensive multilingual view by expanding the variety of programming languages and language-specific error types.}
\resizebox{1.0 \textwidth}{!}{
    \begin{tabular}{l|cccccc}
    \toprule
\bf Benchmark & \bf \#Languages & \bf \#Task & \bf \makecell[c]{Size \\ (Easy/Middle/Hard)} & \bf \makecell[c]{\#Error Types} & \bf \makecell[c]{Source of Bugs} & \bf \makecell[c]{Language-specific \\ Bugs} \\
\midrule 
DeepFix~\citep{yasunaga2021break} & 1  & 1 & 6,971 & 4 & Collection & \textcolor{red}{\ding{55}}\\
Github-Python~\citep{yasunaga2021break} & 1 & 1 & 15K & 14 & Collection & \textcolor{red}{\ding{55}}\\
Bug2Fix~\citep{lu2021codexglue} & 1 & 1 & 5,835 & - & Collection & \textcolor{red}{\ding{55}}\\
FixEval~\citep{haque2023fixeval} & 2 & 1 & 43K/243K &  - & Collection & \textcolor{red}{\ding{55}}\\
CodeError~\citep{wang2023intervenor} & 1 & 1 & 4,463 & 6 & Collection & \textcolor{red}{\ding{55}}\\
CodeEditorBench~\citep{guo2024codeeditorbench} & 3 & 1 & 676/515/716 & 14 & GPT-4 Generation& \textcolor{red}{\ding{55}}\\
DebugBench~\citep{debugbench} & 3 &  1 & 1,438/1,401/1,414 & 18 & GPT-4 Generation & \textcolor{red}{\ding{55}}\\
DebugEval~\citep{debugeval} & 3 & 4  & 1,933/1,903/1,876  & 18 & \makecell[c]{Collection \& GPT-4 Generation} & \textcolor{red}{\ding{55}} \\
\midrule
\benchmark{}~(Ours) & 20 & 3 & 1,692/1,209/612 & 47  & Human Annotation & \textcolor{green}{\ding{51}}\\ 
\bottomrule
\end{tabular}
}
\label{tab:compare_bench}
\end{table*}

\subsection{\benchmark{} Construction \& Quality Control}
To curate the massively multilingual code debugging evaluation benchmark \benchmark{}, we employ a comprehensive and systematic human annotation process for multilingual code samples. This process is guided by meticulously defined guidelines to guarantee accuracy and consistency.

We initially recruite 13 computer science graduates as multilingual debugging annotators, all proficient in their respective programming languages. After completing a comprehensive training course on annotation methods, the annotators are tasked with defining problems, providing corresponding solutions, and buggy code. Annotators adhere to the following principles: (1) Write a clear problem question and design test cases to ensure that bugs can be effectively identified; (2) Categorize bugs into multiple difficulty levels (easy/medium/hard) based on the complexity of fixing these code.

\begin{figure*}[h]
\begin{center}
    \includegraphics[width=1.0\textwidth]{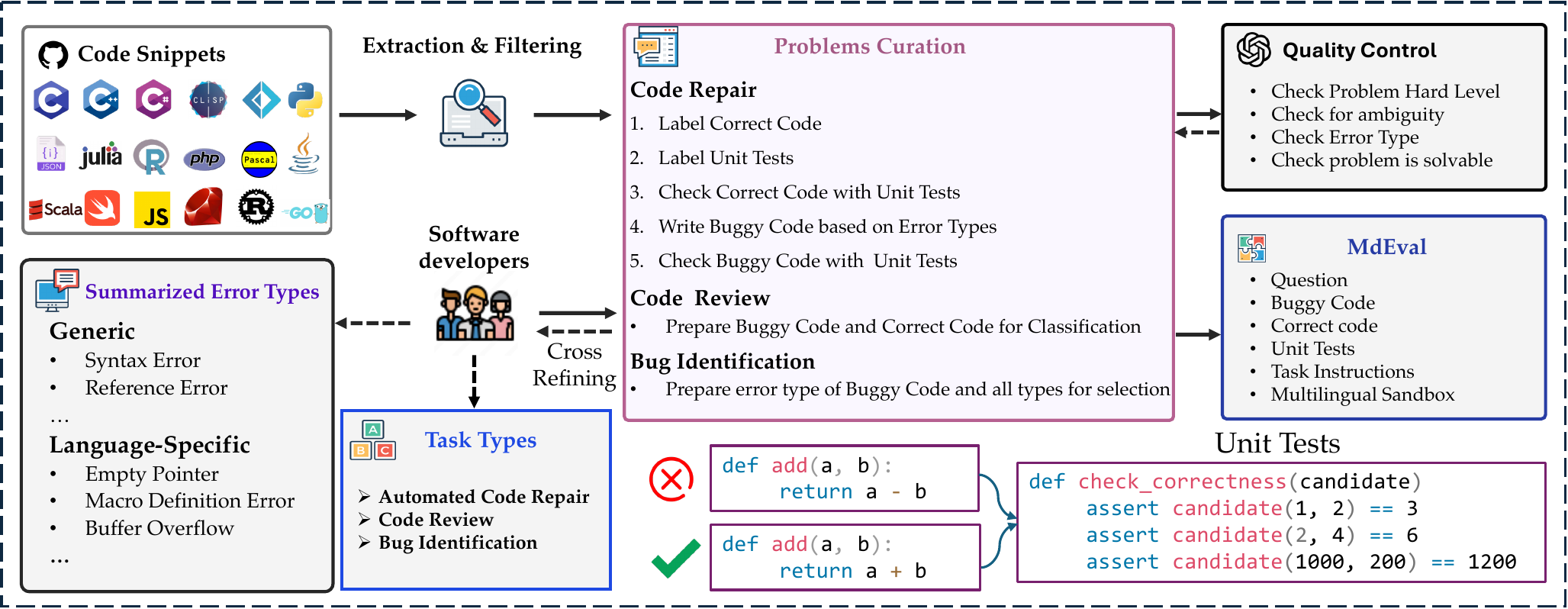}
    \vspace{-5pt}
    \caption{Overview of the \benchmark{} construction process. We collected and filtered code snippets from GitHub. Before annotation, we summarized error types. Annotators then labeled the code based on these types. To ensure quality, they used GPT-4o to evaluate the annotations on four criteria: difficulty, ambiguity, error type, and solvable. Finally, they exchanged data with each other to minimize bias and errors.}
    \label{fig:bench_construct}
    \vspace{-10pt}
\end{center}
\end{figure*}

\autoref{fig:bench_construct} illustrates the overall process of dataset construction. We begin by collecting code snippets from GitHub, which are then extracted and filtered following StarCoder~\citep{starcoder}. Prior to the annotation phase, we summarize generic error types and language-specific error types. The three task definitions and corresponding annotation methods are explained in detail. The annotators proceede to annotate the code according to the identified error types and specified annotation methods. To ensure annotation quality, they evaluate the annotated code based on four criteria: problem difficulty, ambiguity, error type, and solvability. Furthermore, after completing their annotations, each annotator exchanges data with another annotator for cross-refining, aiming to minimize subjective bias and errors. Any discrepancies between annotators are resolved through consensus or with input from senior annotators. Finally, we engage three volunteers to assess the accuracy of the benchmark (targeting > 90\%) and correct errors.

\begin{figure*}[t!]
\begin{center}
    \includegraphics[width=1.0\textwidth]{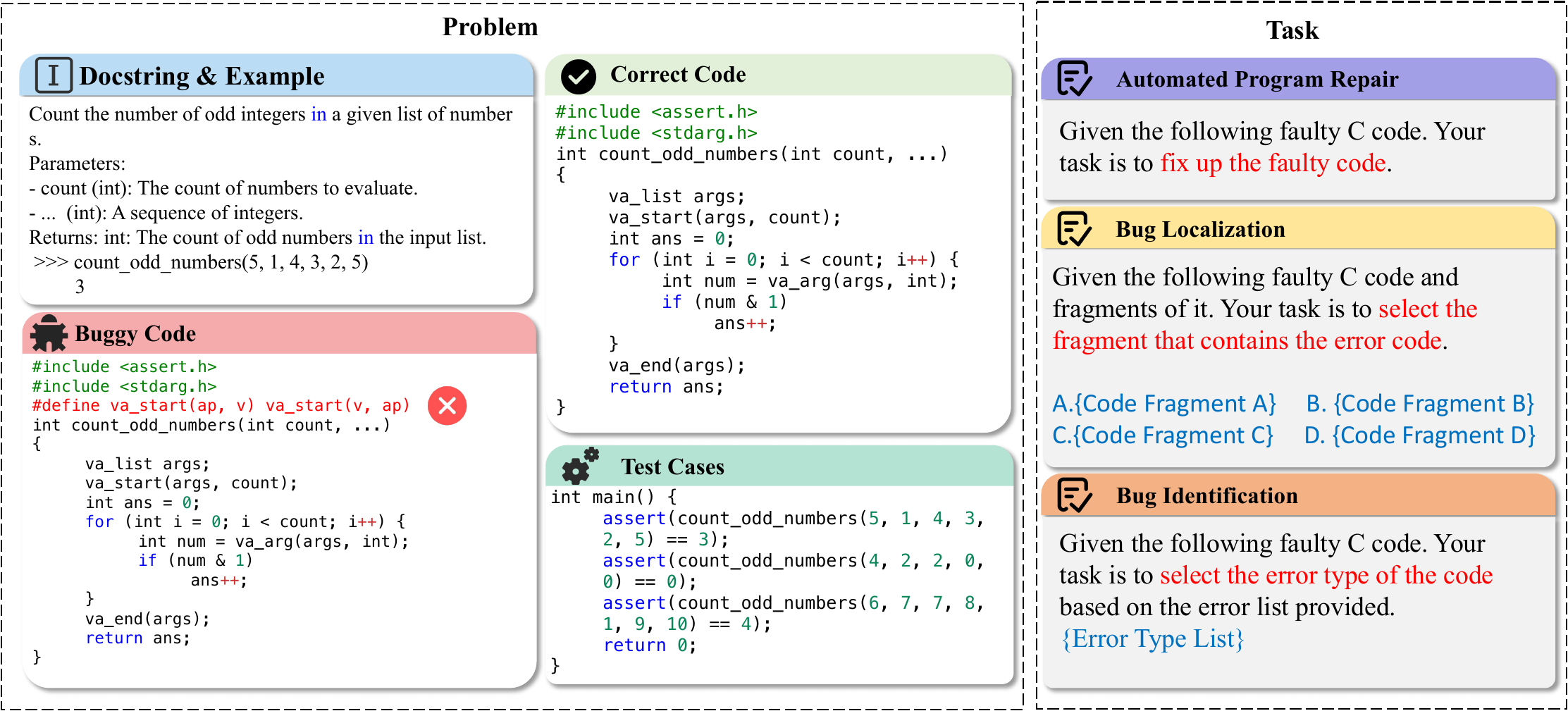}
    \caption{Examples of multilingual automated program repair (APR), bug localization (BL), and code identification (CI). }
    \label{fig:bench_cases}
    \vspace{-10pt}
\end{center}
\end{figure*}

\subsection{Instruction Corpora for Code Debugging}
To create the instruction corpora, we need to create the pair of the correct code snippet and the buggy code. First, we select the proper code snippet from 20 languages and prompt the code LLM to generate a new question $q^{L_{k}}$ of programming language $L_{k}$. Then, we use the LLM to generate the correct code $c^{L_{k}}$ and filter the low-quality response with an LLM filter and the generated test cases. Therefore, we can regard the $(q^{L_{k}}, c^{L_{k}})$ as the correct sample by ensuring the correctness of $c^{L_{k}}$ as much as possible. We propose \ourmethod{} comprised of the following three strategies to create the code debugging instruction corpora \instruct{} to obtain the fine-tuned LLM \baseline{}.

\paragraph{Injecting Bugs into Query.} We can prompt the LLM to modify the original question to another similar question with minor differences, where the similar question $q'^{L_{k}}$ is used to generate the answer $\mathcal{M}_{w}(q'^{L_{k}})$ by a weak LLM $\mathcal{M}_{w}$ with small size (e.g. Qwen2.5-1.5B). Since there exist differences between the original question $q^{L_{k}}$ and the modified question $q'^{L_{k}}$, $(\mathcal{M}_{w}(q'^{L_{k}}), c^{L_{k}})$ can be fed into the LLM as source input and target prediction.

\paragraph{Injecting Bugs into Solution.} Another more intuitive method is to directly inject the bugs into the correct code $c^{L_{k}}$. Given the bug type and the correct code snippet, we prompt the LLM to generate the buggy code $\mathcal{M}(c^{L_{k}})$. The pair $(\mathcal{M}(c^{L_{k}}),c^{L_{k}})$ can be used for the instruction tuning.

\paragraph{Injecting Bugs with Round-trip Code Translation.} Under the multilingual scenario, we can translate the correct $c^{L_{k}}$ into the $\mathcal{M}_{w}(c^{L_{k}}; L_{k} \to L_{j})$ and then back-translate into the original language $L_{k}$ of programming languages using the weak LLM $\mathcal{M}$, where the round-trip translation code snippet can be regarded as the buggy code. The pair $(\mathcal{M}_{w}(\mathcal{M}_{w}(c^{L_{k}}; L_{k} \to L_{j}); L_{j} \to L_{k}),c^{L_{k}})$ can be used for the instruction tuning.

\subsection{Evaluation Task}
\paragraph{Automated Program Repair (APR).}
The automated program repair task forces the LLM to fix the bug in the given code snippet and then generates the correct code. Given the programming language $L_{k} \in \{L_{i}\}_{i=1}^{K}$ ($K=20$ is the number of programming languages), we provide the question $q^{L_{k}}$, the corresponding buggy code $b^{L_{k}}$, and the examples test cases $e^{L_{k}}$ for inputs. We can organize the different input settings for evaluation:
\begin{MiddleEquation}
\begin{align}
    r^{L_{k}} = \mathbb{I}(P(c^{L_{k}}|I;\mathcal{M}); u^{L_{k}})
    \label{eval_code_apr}
\end{align}
\end{MiddleEquation}where $\mathbb{I}(\cdot)$ is the executor of the multilingual sandbox to verify the correctness of the generated code with the test cases $u^{L_{k}}$ (If the fixed code $c^{L_{k}}$ passes all test cases, the evaluation result $r^{L_{k}}=1$, else $0$). In our work, we provide three settings for evaluation to simulate the realistic user queries: (1) Question with buggy code: $I=\{q^{L_{k}},b^{L_{k}}\}$ (2) Buggy code with example test cases: $I=\{b^{L_{k}},e^{L_{k}}\}$ (3) Only buggy code: $I=\{b^{L_{k}} \}$.

\paragraph{Bug Localization (BL).} The Bug Localization (BL) task aims to identify the specific line(s) of code within a given buggy program $c^{L_{K}}$ that contains the error. For each test instance in the BL task, a buggy code $c^{L_{K}}$ is provided, from which four code snippets, {$S_A$, $S_B$, $S_C$, $S_D$}, are extracted. The LLMs are then tasked with identifying the golden snippet $S_G$, which contains the error.

\paragraph{Bug Identification (BI).}
In this task, LLMs are required to classify the type of error present in a given buggy program $c^{L_{k}}$ with one error. The LLMs must choose the correct error category from 47 bug types (including generic bug types and language-specific bug types).

\subsection{Evaluation Metrics}
\paragraph{Automated Program Repair.}
In the automated program repair task, we evaluate models by executing the generated code against a set of unit tests and assessing performance using the Pass@1 metric (pass rate for just one-time generation). Greedy Pass@1 indicates whether a result produced by the LLM successfully passes corresponding unit tests. 

\paragraph{Bug Localization \& Bug Identification.}
In the bug localization and bug identification tasks, we evaluate model performance using accuracy, as both require the model to select from a set of provided options.

\section{Experiments}
\subsection{Experiment Setup}
\label{sec:experimental_setup}

\paragraph{Code LLMs.}
We evaluate 40 popular models, including GPTs~\citep{gpt4}, Claude-3.5~\citep{claude}, and code-specific models like Qwen2.5-Coder~\citep{qwencoder}, DeepSeekCoder~\citep{deepseek_coder}, CodeLlama~\citep{code_llama}, and Codegemma~\citep{codegemma}. Additionally, we fine-tune Qwen2.5-Coder-7B as our baseline \baseline{}. 

\paragraph{\baseline{} Training Setup}
The training data for \baseline{} comprises our debugging dataset \instruct{} and the Magicoder-Instruct code generation dataset \citep{magicoder}, ensuring fundamental instruction-following capabilities for code-related tasks. 
\baseline{}, built on Qwen2.5-Coder-7B, is trained for 3 epochs using a cosine scheduler with an initial learning rate of \(5 \times 10^{-5}\) with a 3\% warmup ratio. We employ AdamW~\citep{adamw} as the optimizer, with a batch size of 1024 and a maximum sequence length of 2048. (Details can be found in the Appendix). 

\input{table/code_repair}
\input{table/code_ident}
\input{table/code_loc}

\subsection{Main Results}
\paragraph{Automated Program Repair.}
\autoref{tab:task_repair1} presents the Pass@1 results of different models on \benchmark{} for the multilingual automated program repair task~(given question with buggy code, request the model to fix buggy code). The results indicate a marked disparity between closed-source state-of-the-art models and the majority of open-source models across nearly all programming languages. Notably, GPT-4o, Claude-3.5-sonnet, and Qwen2.5-Coder-Instruct excel in this task and demonstrate significant performance advantages over other models.
Furthermore, our baseline model \baseline{}, is fine-tuned using only 16K bug-related data \instruct{}. Despite the limited size of this dataset, the model demonstrated competitive performance compared to others of similar scale, highlighting the effectiveness of \instruct{} in enhancing the debugging capabilities of models.

\paragraph{Bug Localization}
\autoref{tab:task_loc} illustrates the accuracy of different models on the multilingual bug localization task. It is evident that closed-source models outperform open-source models by a significant margin, demonstrating the superior bug localization capabilities of closed-source models.
Specifically, open-source models with smaller parameter sizes like OpenCoder-1.5B-Instrcut, due to their poor instruction-following capabilities, are unable to output the correct format as required, resulting in lower accuracy in localization.
Besides, it is observed that for the same model, the bug localization accuracy is lower than its pass@1 scores in automated program repair task. We hypothesize that this discrepancy arises because the bug localization task requires a strong understanding of location information, which happens to be a weakness of large language models. Therefore, improving LLMs' ability to understand location information is a critical issue that needs to be addressed.

\paragraph{Bug Identification.}
In the bug identification task, the goal is to identify the error type in a given code snippet, where the LLMs analyze source code for defects and choose the correct bug type from the pre-defined 47 bug types. \autoref{tab:ident} lists the all results of the bug identification. Notably, the closed-source LLMs, such as GPT-4o and Claude series, have the dominant advantages, outperforming the open-source LLMs by nearly +10 points. Bug identification with 47 bug types poses a daunting challenge to the LLMs, requiring alignment capability of LLMs between the given code snippet and its corresponding bug type. As a result, some open-source models with smaller parameter sizes perform poorly in this task

\section{Further Analysis}
\subsection{Performance across error types.}
\begin{wrapfigure}{r}{0.45\linewidth} 
    \centering
    \includegraphics[width=0.45\textwidth]{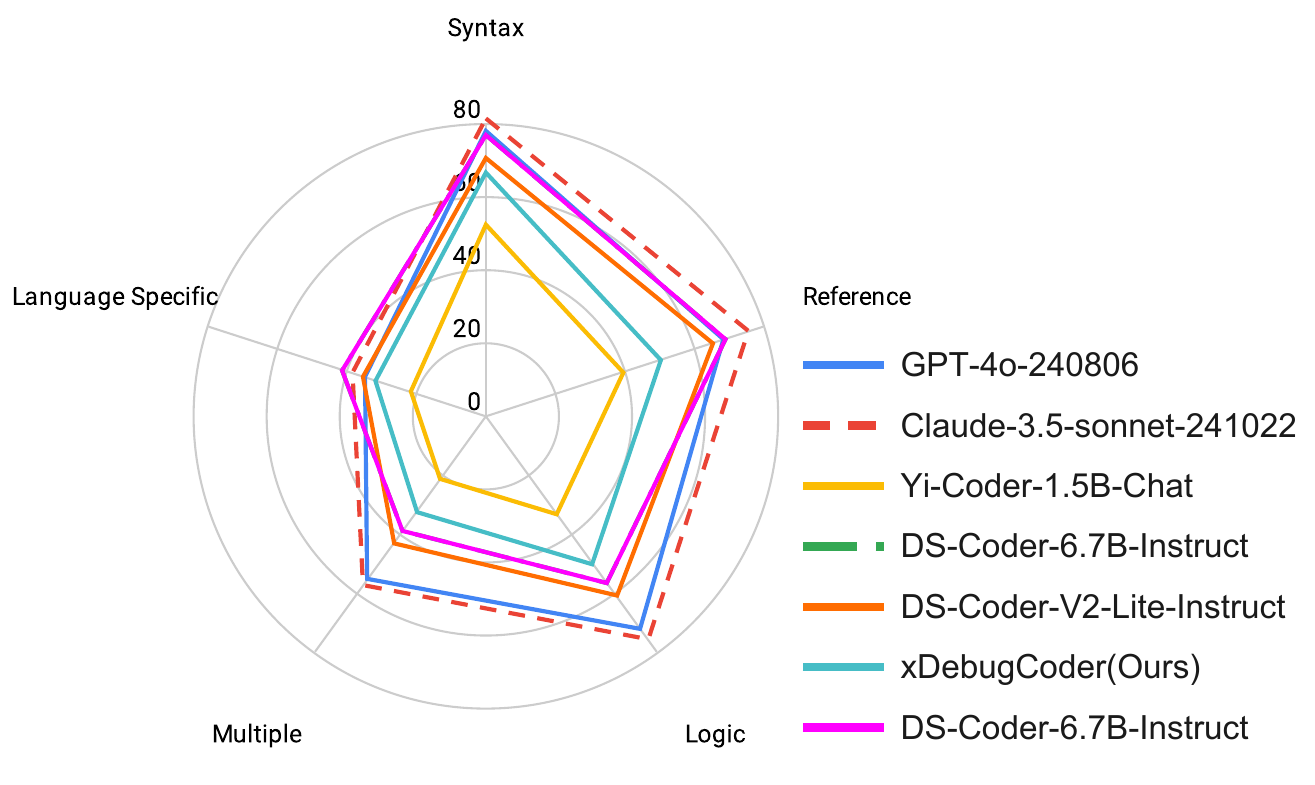}
    \caption{The performance of models on the automated program repair task across error types. }
    \label{fig:repair_error_types}
\end{wrapfigure}
In \autoref{fig:repair_error_types}, The performance of models on the automated program repair task varies across different error types, highlighting the strengths and weaknesses of these models in addressing specific challenges. Consistently, the models demonstrate robust capabilities in repairing syntax errors, reference errors, and logic errors. These error types tend to be more straightforward and well-defined, allowing the models to leverage their knowledge effectively to identify and correct issues with high accuracy.
In contrast, the models exhibit their worst performance when dealing with language-specific errors. 
Language-specific errors can arise from unique syntax rules, idiomatic expressions, or even cultural programming practices that are not universally applicable. As a result, addressing these types of errors presents a significant challenge and underscores the need for further improvements in model training.

\subsection{Another Two Automated Program Repair Settings}
\begin{figure*}[h]
	\centering
	\subfigure[Buggy code with example test cases]{\includegraphics[width=1.0\textwidth]{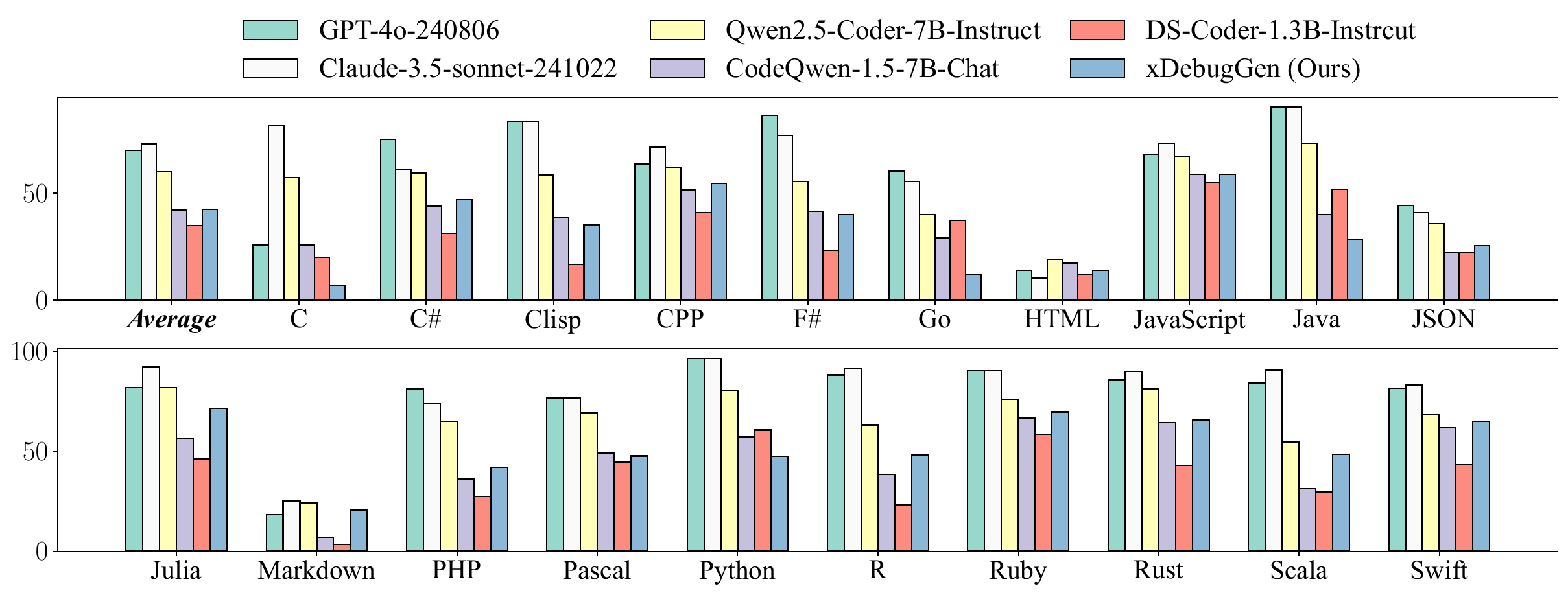}}
	\subfigure[Only buggy code]{\includegraphics[width=1.0\textwidth]{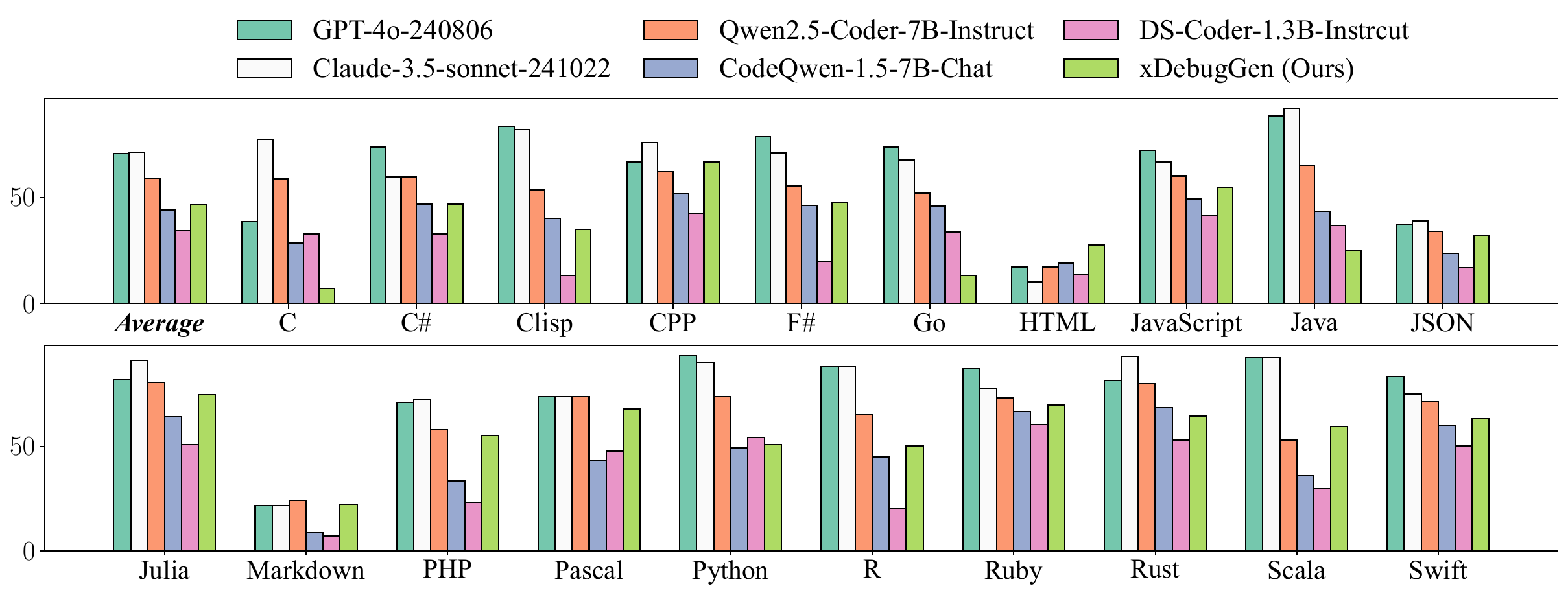}}
	\caption{Two additional automated program repair settings are designed to simulate realistic user queries. Part (a) presents results for the scenario where models are provided with buggy code along with example test cases, while Part (b) illustrates results for the scenario where only the buggy code is provided to the models.}
	\label{fig:repair_setting23}
\end{figure*}

In \autoref{fig:repair_setting23}, we explore two additional automated program repair settings that aim to simulate realistic user queries in software debugging. Part (a) presents the results for the scenario in which models are given both buggy code and corresponding example test cases. This setup allows for a comprehensive evaluation of the ability of models to understand and correct specific issues based on contextual examples. In contrast, Part (b) illustrates the results for a more challenging scenario where only the buggy code is provided to the models, requiring them to identify and rectify errors without any additional context. This comparison highlights the varying capabilities of models in different settings, emphasizing the importance of context in automated program repair.

\subsection{Anslysis of Code Review task}
Besides automated program repair tasks, code review tasks also play a crucial role in software development. To analyze the performance of different models on code review tasks, we conducted experiments based on \benchmark.  For the code review task, we present two versions of code to LLMs: the correct code $b^{L_{k}}$ and the buggy code $c^{L_{k}}$ with only a few minor differences between them. The correct code and buggy code are listed in a random order to feed into LLM for distinguishing the buggy code.

\begin{figure*}[t!]
\begin{center}
    \includegraphics[width=1.0\textwidth]{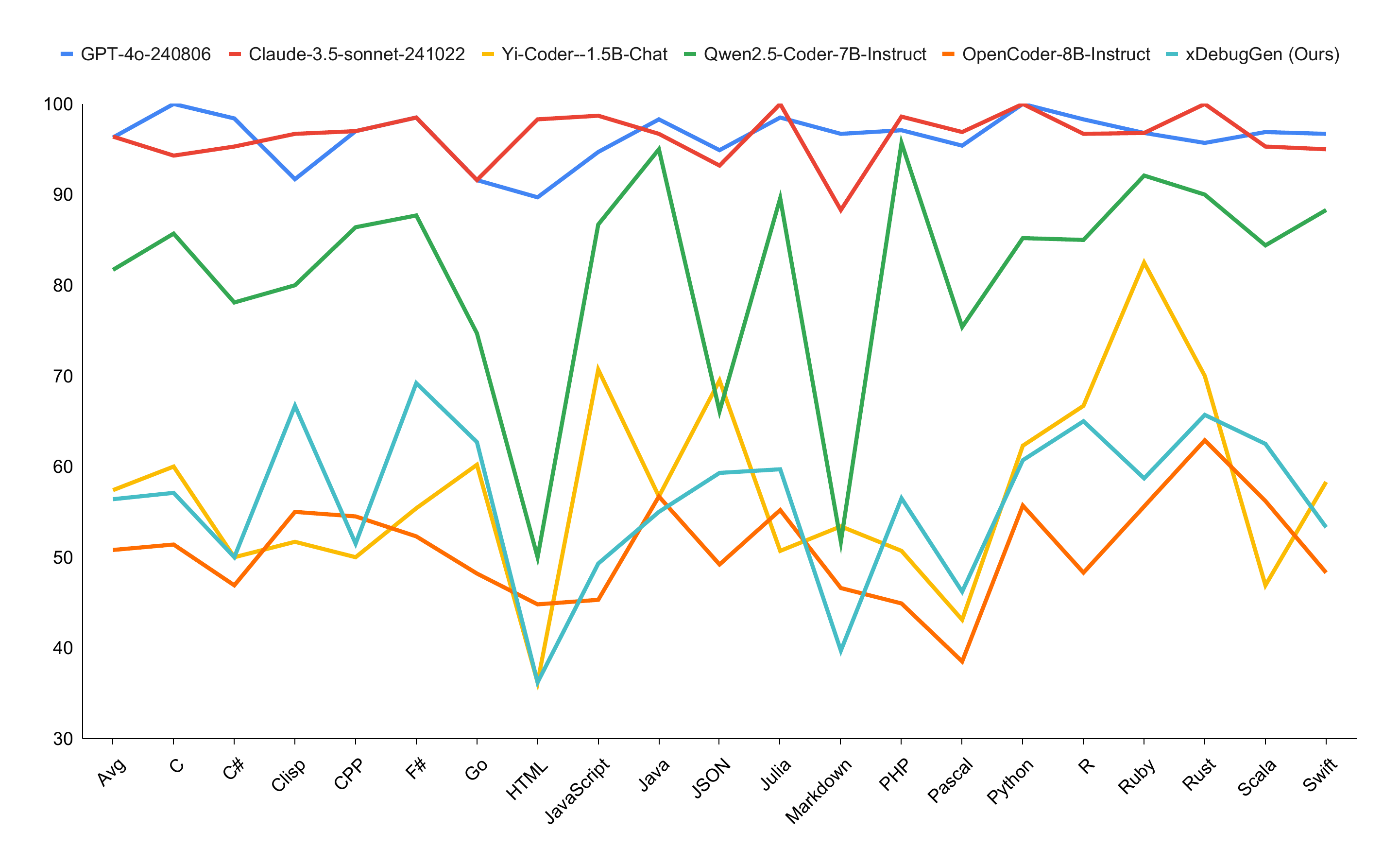}
    \caption{Accuracy of different models for Code Review tasks on \benchmark{}. }
    \label{fig:review}
    \vspace{-10pt}
\end{center}
\end{figure*}

\autoref{fig:review} 
displays the accuracy for code review tasks. The results show that closed-source models still significantly outperform open-source models in the code review task. The closed-source models demonstrate a strong ability to understand complex code logic, achieving an accuracy rate of approximately 90\%. In contrast, the smaller open-source model exhibits significant challenges, with an accuracy rate of around 50\%. This disparity underscores the limitations of the current open-source model in effectively interpreting intricate coding patterns.

\subsection{Analysis on the Effect of Bug Location for Automated Program Repair}
\begin{figure*}[t!]
\begin{center}
    \includegraphics[width=1.0\textwidth]{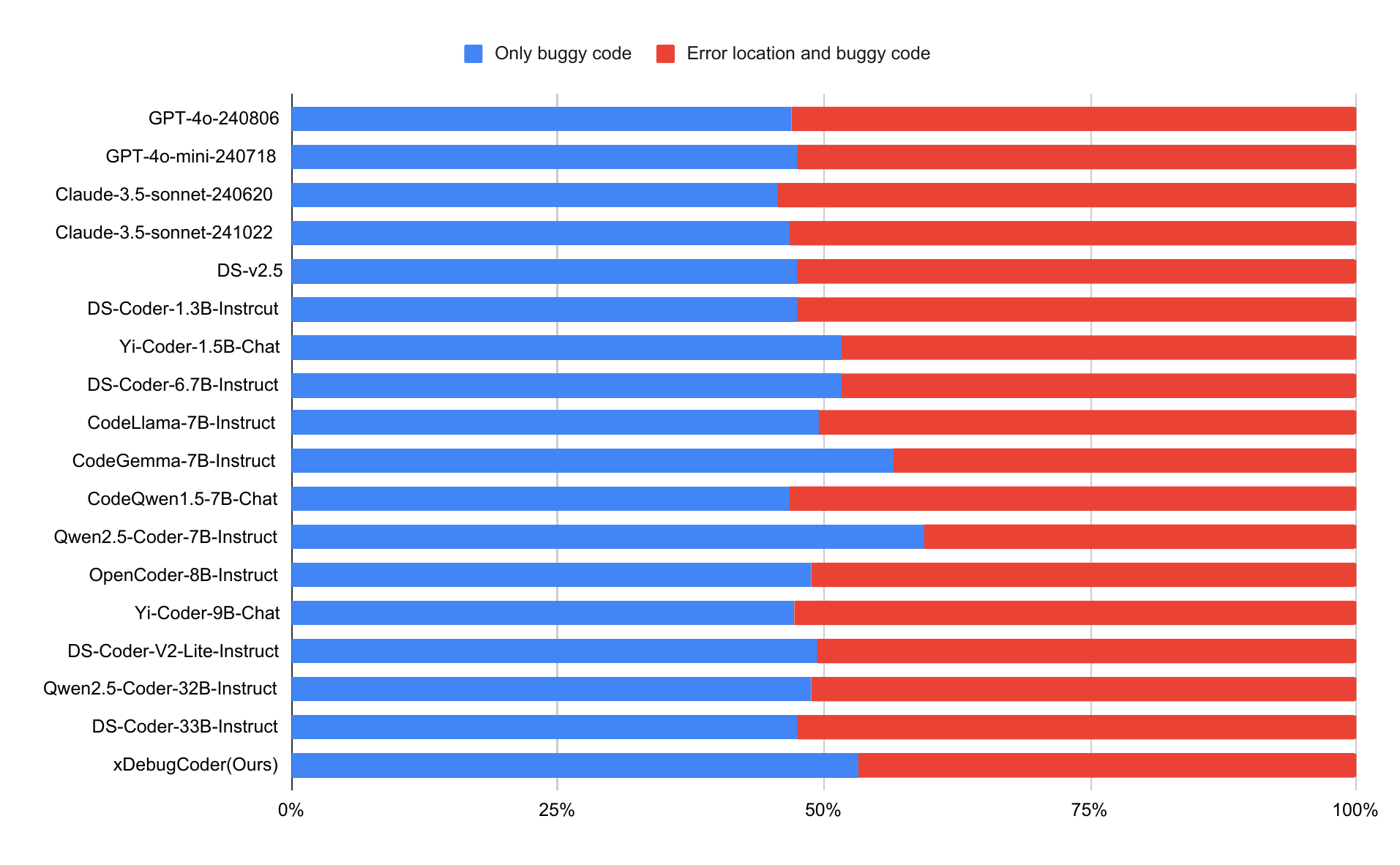}
    \caption{Comparison of the Pass@1 (\%) scores  with only the buggy code provided versus when additional bug location information is supplied.}
    \label{fig:loc_apr}
    \vspace{-10pt}
\end{center}
\end{figure*}

In previous studies, bug localization has been regarded as the first step in program repair, playing a critical role. To verify whether the bug location information can also have a positive impact when using large language models for automated program repair, we designed and conducted a series of comparative experiments, as shown in ~\autoref{fig:loc_apr}.
We test two scenarios: providing the bug location information and not providing it and task the model with repairing buggy code in both cases. The results indicate that providing the bug location information significantly improves Pass@1 scores of automated program repair.
However, our prior experimental results reveal that for LLMs, the difficulty of the bug localization task is notably higher than that of the automated program repair task. Therefore, improving the bug localization capabilities of the model is essential for enhancing its overall automated program repair performance.

\section{Related Work}
\label{sec:related_work}

\paragraph{Code Large Language Model.} 
The rapid progress of large language models\citep{gpt4,llama2,llama3,Qwen,qwen2} has enabled complex code-related tasks. Early models like BERT\citep{bert} and GPT~\citep{gpt}, trained on billions of code snippets, focused on code understanding and generation~\citep{codex, code_bert, bloom, AlphaCode,codet5,santacoder}. Recent advances in domain-specific pre-training and instruction fine-tuning~\citep{kun,mammoth2} have enhanced models like CodeLlama~\citep{code_llama} and WizardCoder~\citep{wizardcoder}, achieving strong performance in code completion, synthesis, and repair.

\paragraph{Debugging with Large Language Models.} 
LLMs have also gained popularity for automatic program debugging, a critical task for bug detection, vulnerability identification~\citep{pradel2018deepbugs,allamanis2021self}, fuzz testing~\citep{deng2023large,xia2024fuzz4all}, and program repair~\citep{wen2024fixing,gu2024counterfeit}. Benchmark tests, such as DebugBench~\citep{debugbench} and DebugEval~\citep{debugeval}, assess LLM debugging capabilities across error categories and tasks. However, these focus on 1-3 languages, neglecting language-specific errors. To fill this gap, we propose \benchmark{}, a comprehensive debugging benchmark for 20 languages to evaluate LLM performance from a broader perspective.

\section{Conclusion}
\label{sec:conclusion}
In this work, we introduce \benchmark{} of instruction corpora \instruct{}, evaluation benchmark, and a strong baseline \baseline{}, where the benchmark includes automated program repair (APR), bug localization (BL), and bug identification (BI) of 20 programming languages (total 3.9K samples), aiming to assess the debugging capabilities of large language models (LLMs) in multilingual environments. Further, we propose \ourmethod{} to construct a multilingual debugging instruction corpus, where we inject the bugs into the query or answer to create the pair of the buggy code and correct code. Based on \instruct{}, we develop \baseline{}, a multilingual LLM for debugging in a wide range of programming languages as a strong baseline. Through extensive experiments, this paper reveals a substantial performance gap between open-source and closed-source LLMs, underscoring the need for further improvements in multilingual code debugging. In the future, we will continue expanding the number of languages in \benchmark{}.


\bibliography{reference}
\bibliographystyle{natbib}

\clearpage
\appendix

\section{Limitations}
\paragraph{Language Coverage.} Although \benchmark{} covers 20 programming languages, there are still many languages not included, particularly those that are less commonly used or have niche applications. Expanding the benchmark to include more languages would provide a more comprehensive evaluation of multilingual debugging capabilities.

\paragraph{Real-world Applicability.} While \benchmark{} aims to simulate realistic debugging scenarios, the tasks and data may not fully capture the complexity and variability of real-world software development. Incorporating more diverse and complex real-world projects into the benchmark could improve its applicability and relevance.

\paragraph{Instruction Tuning Data.} The instruction corpora \instruct{} used for fine-tuning the baseline model \baseline{} is generated by LLM-based bug injection. While this approach has shown promise, the quality and diversity of the generated data could be further improved. Exploring alternative methods for generating high-quality instruction data, such as leveraging more advanced LLMs or incorporating feedback from real-world debugging sessions, could enhance the effectiveness of the instruction tuning process.

\section{Ethical Considerations}
\subsection{Potential Risks}
\benchmark{}, as evaluation tools, can comprehensively assess the capability of large language models in debugging tasks across a wide range of programming languages, thereby advancing the development of large language models in this domain. However, improper or erroneous use of \benchmark{} may pose significant risks, such as incorrect program analysis and faulty program repair, which could even lead to severe consequences such as program crashes or operating system failures. Therefore, to ensure the security and reliability of the evaluation process, we strongly recommend using \benchmark{} within a sandbox environment. Such an environment can effectively isolate potential system risks, ensuring the accuracy and safety of the evaluation.

\section{Human Annotation}

To construct the massively multilingual code debugging benchmark \benchmark{}, we designed and implemented a comprehensive and systematic human annotation process to ensure the accuracy, consistency, and high quality of multilingual code samples. This process strictly adheres to carefully formulated annotation guidelines and incorporates multiple quality control mechanisms.

We recruited 13 computer science graduates as multilingual debugging annotators, all of whom are proficient in at least one programming language and possess a solid foundation in computer science. Prior to the formal annotation process, annotators underwent systematic training on annotation methods, covering core tasks such as problem definition, solution design, and buggy code generation.

Our annotation training guidelines focus on the following key aspects:
\begin{itemize}
\item \textbf{Standardized Format}: We provide detailed annotation examples and templates for 20 programming languages. Annotators must strictly adhere to a standardized format throughout the annotation process to ensure data consistency and reusability.
\item \textbf{Accessibility}: All annotation reference data are sourced from open-source materials that allow free use and distribution, ensuring compliance with academic research purposes and relevant legal and ethical requirements.
\item \textbf{Difficulty Classification}: We establish a detailed difficulty classification guideline for each programming language. Annotators must categorize each problem according to complexity, error type, and problem scale, assigning an appropriate difficulty level (e.g., easy, middle, hard) following the guidelines.
\item \textbf{Self-Containment}: Annotators must ensure that each problem description is complete and unambiguous, containing all necessary information for problem-solving. Provided example inputs and outputs must be accurate, the generated buggy code must be ensured to fail execution correctly, and the reference solution must pass all test cases. Additionally, test cases should comprehensively cover various boundary conditions and exceptional scenarios.
\end{itemize}

To maintain annotation quality and incentivize annotators, we offered a compensation of approximately \$6 per problem. Moreover, we provided annotators with a comfortable working environment, free meals, souvenirs, and high-performance computing equipment. A total of approximately 1,300 problems were annotated, with additional annotators hired for quality inspection, leading to a total cost of around \$5,000. Quality inspection tasks included bug identification, bug localization, and code review.

\subsection{Quality Control}

To ensure the high quality of the \benchmark{}, we implemented a rigorous quality control mechanism. First, annotators were required to evaluate the annotated code based on four core criteria: problem difficulty, ambiguity, error type, and solvability. Second, we adopted a dual verification system, where each code snippet was independently annotated by at least two annotators to minimize subjective bias and human errors. In cases of disagreement, resolution was achieved through discussion or by a senior annotator making the final decision.

To further ensure the reliability of the benchmark, we employed three volunteers to assess whether \benchmark{} achieved a correctness rate of at least 90\% and to correct any errors, thereby guaranteeing the accuracy of the annotations.

\section{Experiment Detail}
\paragraph{\baseline{} Training Corpora.}
The training corpora consist of our debugging dataset~\instruct{}, which contains 16K samples, and the Magicoder-Instruct code generation dataset \citep{magicoder}, comprising 180K samples. This combination ensures that the model possesses a fundamental capability to follow instructions for basic code tasks.  We apply data decontamination before training our \ourmethod{}. Following \citet{starcoder,magicoder}, we adopt the N-gram exact match decontamination method with \benchmark{}, HumanEval~\citep{codex}, MultiPL-E~\citep{multipl_e}, MBPP~\citep{mbpp}.

\paragraph{\baseline{} Optimization.}
Our model, \baseline{}, based on Qwen2.5-Coder-7B, is trained for 3 epochs using a cosine scheduler, starting at a learning rate of \(5 \times 10^{-5}\) with 3\% of total training steps for warmup. We utilize AdamW~\citep{adamw} as the optimizer; the batch size is set to 1024, with a maximum sequence length of 2048. 
All experiments are performed with 8 NVIDIA A800-80GB GPUs.

\paragraph{Code LLMs.} 
We evaluate 40 popular models, both closed-source and open-source (sizes ranging from 1.3B to 605B parameters).
For general models, we evaluate GPTs~\citep{gpt4} (GPT4-o, GPT4-o-mini), Claude-3.5~\citep{claude}.
For code models, we test Qwen2.5-Coder~\citep{qwencoder}, DeepSeekCoder~(DS-Coder)~\citep{deepseek_coder}, CodeLlama~\citep{code_llama}, and Codegemma~\citep{codegemma}.
Furthermore, we fine-tune the Qwen2.5-Coder-7B to provide a baseline model \baseline{} for reference. For closed-source models, the responses are generated by the official API. For the open-source models, we perform inference on all models using the vLLM~\citep{vllm} framework. All models adopt a greedy decoding strategy during inference, the temperature is set to 0, and the maximum generation length is 4096. 

\section{Related Work}
\paragraph{Code Large Language Model.}
With the rapid advancement of large language models(LLMs)~\citep{gpt4,llama2,llama3,Qwen,qwen2}, solving complex code-related tasks has become increasingly feasible, leading to the emergence of numerous Code LLMs.
Early studies utilized models like BERT~\citep{bert} or GPT~\citep{gpt} as backbones, trained on billions of code snippets to enable tasks involving code understanding and generation~\citep{codex, code_bert, bloom, AlphaCode,codet5,santacoder}. 
Recently, advancements in domain-specific pre-training and instruction fine-tuning techniques~\citep{kun,mammoth2} have led to extensive efforts in fine-tuning models on large-scale code corpora and crafting code-related task instructions~\citep{code_llama,codegeex,wizardcoder,octopack,codegemma,opencodeinterpreter,deepseek_coder,magicoder,unicoder,starcoder2,mistral,qwencoder,mac_sql,r2c2coder,m2rceval}. These models demonstrate remarkable performance in tasks like code completion, synthesis, and program repair.

\paragraph{Debugging with Large Language Models.}
Automatic program debugging holds substantial practical value. With the emergence of LLM capabilities, a growing number of individuals are utilizing LLMs for code debugging, leading to extensive research in this field.
Code Debugging includes serval tasks such as bug or vulnerability detection~\citep{pradel2018deepbugs,allamanis2021self,yuan2023evaluating,zhang2024prompt,zhong2024advancing}, fuzz test~\citep{deng2023large,xia2024fuzz4all,yang2024fuzzcoder}, program repair ~\citep{wen2024fixing,quixbugs,EvalGPTFix,runbugrun,gu2024counterfeit,tambon2024bugs,wang2024large}, GitHub issues auto resolving ~\citep{swe_bench,coder,magis}.
To effectively assess the code debugging capabilities of LLMs, several benchmark tests have been introduced ~\citep{prenner2022can,sobania2023analysis,xia2023conversational,EvalGPTFix,debugbench,debugeval}. Notably, DebugBench~\citep{debugbench} provides a comprehensive classification of error types and analyzes the debugging capabilities of LLMs based on these categories. Similarly, DebugEval~\citep{debugeval} has designed various debugging-related tasks to evaluate LLM performance across different task dimensions. However, these studies focus on 1 to 3 languages. In reality, there are significant differences in code errors between languages, leading to numerous language-specific errors. To address this gap, we propose \benchmark{}, a comprehensive code debugging benchmark covering 20 languages, aiming to assess LLM debugging capabilities from a broader perspective.

\end{document}

%% file: table/code_repair.tex
\begin{table*}[t]
  \centering
   \caption{Pass@1 (\%) scores of different models for Automated Program Repair tasks on \benchmark{}. The underlined numbers are the best scores for each language. ``Avg$_{all}$'' represents the average scores of all code languages.}
  \resizebox{1.0\textwidth}{!}{
  \begin{tabular}{l|cccccccccccccccccccccc}
  \toprule
\textbf{Model} &  Size & \textbf{Avg$_{all}$}  & C & C\# & CLISP & CPP & F\# & Go &HTML  & JS & Java  & Json   & Julia & MD & PHP & Pascal & Python & R & Ruby & Rust & Scala & Swift  \\ \midrule
        \multicolumn{23}{c}{\textbf{Closed-Source Models}} \\
        \midrule
        o1-preview & \faLock{} & 70.2 & 68.6 & 73.1 & \underline{91.7} & 63.8 & \underline{89.2} & 38.6 & 15.5 & \underline{84.0} & 90.0 & 39.0 & 89.6 & 20.0 & 84.1 & \underline{80.0} & 91.8 & 60.0 & 93.7 & 85.7 & 84.4 & 56.7 \\ 
        o1-mini & \faLock{} & \underline{72.9} & 65.7 & \underline{76.1} & 60.0 & \underline{68.1} & 81.5 & \underline{68.7} & 5.2 & 80.0 & 91.7 & 42.4 & \underline{92.5} & \underline{25.0} & \underline{87.0} & 78.5 & 90.2 & \underline{88.3} & \underline{96.8} & \underline{98.6} & 87.5 & 60.0 \\ 
        GPT-4o-240806 & \faLock{} & 67.5 & 14.3 & 64.1 & 85.0 & 66.7 & 86.2 & 56.6 & 13.8 & 57.3 & 83.3 & 42.4 & 80.6 & 20.0 & 87.0 & 72.3 & 91.8 & 86.7 & 84.1 & 84.3 & 89.1 & 86.7 \\ 
        GPT-4o-mini-240718 & \faLock{} & 65.3 & 18.6 & 64.1 & 71.7 & 57.6 & 75.4 & 56.6 & 8.6 & 61.3 & 85.0 & \underline{47.5} & 83.6 & 23.3 & 85.5 & 67.7 & 88.5 & 80.0 & 81.0 & 87.1 & 76.6 & 85.0 \\ 
        GPT-4-Turbo-240409 & \faLock{} & 61.7 & 24.3 & 53.7 & 63.3 & 49.3 & 84.6 & 50.6 & 3.4 & 60.0 & 80.0 & 35.6 & 74.6 & 21.7 & 81.2 & 75.4 & \underline{95.0} & 76.7 & 82.5 & 81.4 & 81.2 & 56.7 \\ 
        Claude-3.5-sonnet-240620 & \faLock{} & 66.0 & 34.3 & 56.2 & 83.3 & 60.6 & 83.1 & 63.9 & 5.2 & 65.3 & 70.0 & \underline{47.5} & 68.7 & 20.0 & 76.8 & 67.7 & 91.8 & 71.7 & 84.1 & 90.0 & \underline{93.8} & 80.0 \\ 
        Claude-3.5-sonnet-241022 & \faLock{} & 70.3 & \underline{81.4} & 57.8 & 86.7 & 59.1 & \underline{89.2} & 44.6 & 8.6 & 60.0 & 91.7 & 44.1 & 82.1 & 21.7 & 82.6 & 75.4 & 82.0 & 80.0 & 85.7 & 88.6 & \underline{93.8} & 90.0 \\ 
        Yi-lighting & \faLock{} & 57.8 & 24.3 & 53.7 & 60.0 & 55.1 & 67.7 & 41.0 & 5.2 & 60.0 & 76.7 & 25.4 & 82.1 & 8.3 & 78.3 & 63.1 & 91.7 & 75.0 & 79.4 & 81.4 & 78.1 & 46.7 \\ 
        Doubao-Pro & \faLock{} & 60.2 & 68.6 & 55.2 & 56.7 & 55.1 & 78.5 & 53.0 & 8.6 & 56.0 & 80.0 & 15.3 & 70.1 & 8.3 & 72.5 & 66.2 & 85.0 & 81.7 & 82.5 & 87.1 & 78.1 & 35.0 \\ 
        
        \midrule
        \multicolumn{23}{c}{\textbf{0.5B+ Models}} \\
        \midrule
        
        Qwen2.5-Instruct & 0.5B & 20.6 & 28.6 & 10.4 & 8.3 & 14.5 & 9.2 & 1.2 & 13.8 & 45.3 & 28.3 & 10.2 & 26.9 & 5.0 & 17.4 & 13.8 & 39.3 & 6.7 & 58.7 & 24.3 & 18.8 & 31.7 \\ 
        DS-Coder-Instruct & 1.3B & 33.6 & 28.6 & 42.2 & 13.3 & 43.9 & 24.6 & 38.6 & 5.2 & 44.0 & 48.3 & 18.6 & 47.8 & 1.7 & 33.3 & 27.7 & 44.3 & 16.7 & 61.9 & 41.4 & 34.4 & 45.0 \\ 
        Qwen2.5-Instruct & 1.5B & 35.5 & 24.3 & 32.8 & 15.0 & 27.5 & 23.1 & 18.1 & 8.6 & 60.0 & 50.0 & 28.8 & 55.2 & 8.3 & 34.8 & 30.8 & 62.3 & 20.0 & 69.8 & 67.1 & 32.8 & 35.0 \\ 
        OpenCoder-Instruct & 1.5B & 34.8 & 15.7 & 13.4 & 20.0 & 26.1 & 26.2 & 15.7 & 12.1 & 57.3 & 58.3 & 15.3 & 55.2 & 8.3 & 36.2 & 47.7 & 54.1 & 31.7 & 68.3 & 52.9 & 51.6 & 28.3 \\ 
        Yi-Coder-Chat & 1.5B & 32.4 & 37.1 & 34.4 & 3.3 & 30.3 & 7.7 & 28.9 & 8.6 & 45.3 & 53.3 & 15.3 & 55.2 & 1.7 & 34.8 & 41.5 & 52.5 & 28.3 & 49.2 & 42.9 & 28.1 & 40.0 \\ 
        Qwen2.5-Coder-Instruct & 1.5B & 34.8 & 11.4 & 26.9 & 15.0 & 30.4 & 16.9 & 20.5 & 17.2 & 61.3 & 45.0 & 28.8 & 58.2 & 10.0 & 40.6 & 36.9 & 55.7 & 28.3 & 60.3 & 58.6 & 29.7 & 40.0 \\ 
        Qwen2.5-Instruct & 3B & 46.0 & 51.4 & 40.3 & 28.3 & 36.2 & 41.5 & 41.0 & 12.1 & 69.3 & 61.7 & 27.1 & 61.2 & 11.7 & 53.6 & 47.7 & 63.9 & 45.0 & 58.7 & 65.7 & 53.1 & 38.3 \\ 
 
        \midrule
        \multicolumn{23}{c}{\textbf{6B+ Models}} \\
        \midrule
        
        DS-Coder-Instruct & 6.7B & 56.3 & 37.1 & 60.9 & 56.7 & 63.6 & 60.0 & 56.6 & 8.6 & 61.3 & 75.0 & 23.7 & 64.2 & 6.9 & 52.2 & 60.0 & 78.7 & 51.7 & 88.9 & 80.0 & 60.9 & 68.3 \\ 
        CodeQwen1.5-chat & 7B & 42.6 & 34.3 & 34.4 & 43.3 & 33.3 & 41.5 & 42.2 & 10.3 & 54.7 & 55.0 & 20.3 & 62.7 & 8.6 & 49.3 & 41.5 & 52.5 & 30.0 & 69.8 & 62.9 & 34.4 & 61.7 \\ 
        CodeLlama-Instruct & 7B & 27.2 & 2.9 & 20.3 & 25.0 & 25.8 & 24.6 & 22.9 & 19.0 & 53.3 & 6.7 & 15.3 & 37.3 & 12.1 & 24.6 & 33.8 & 42.6 & 16.7 & 50.8 & 48.6 & 14.1 & 41.7 \\ 
        CodeGemma-Instruct & 7B & 45.9 & 34.3 & 32.8 & 3.3 & 43.9 & 44.6 & 44.6 & 19.0 & 60.0 & 68.3 & 25.4 & 64.2 & 0.0 & 56.5 & 36.9 & 65.6 & 40.0 & 73.0 & 67.1 & 56.2 & 70.0 \\ 
        Qwen2.5-Instruct & 7B & 50.4 & 57.1 & 47.8 & 38.3 & 50.7 & 61.5 & 26.5 & 8.6 & 60.0 & 81.7 & 32.2 & 61.2 & 8.3 & 73.9 & 47.7 & 70.5 & 50.0 & 58.7 & 62.9 & 60.9 & 45.0 \\ 
        Qwen2.5-Coder-Instruct & 7B & 61.7 & 58.6 & 60.9 & 61.7 & 60.6 & 70.8 & 47.0 & 19.0 & 60.0 & 81.7 & 37.3 & 74.6 & 22.4 & 73.9 & 61.5 & 77.0 & 65.0 & 73.0 & 78.6 & 70.3 & 75.0 \\ 
        OpenCoder-Instruct & 8B & 53.4 & 10.0 & 56.2 & 46.7 & 50.0 & 66.2 & 8.4 & 13.8 & 66.7 & 78.3 & 27.1 & 74.6 & 15.5 & 62.3 & 53.8 & 77.0 & 61.7 & 76.2 & 81.4 & 71.9 & 75.0 \\ 
        Meta-Llama-3-Instruct & 8B & 37.9 & 51.4 & 35.8 & 8.3 & 42.0 & 30.8 & 21.7 & 10.3 & 60.0 & 41.7 & 20.3 & 49.3 & 0.0 & 55.1 & 40.0 & 57.4 & 50.0 & 49.2 & 48.6 & 46.9 & 28.3 \\ 
        Meta-Llama-3.1-Instruct & 8B & 42.1 & 57.1 & 41.8 & 26.7 & 40.6 & 44.6 & 21.7 & 6.9 & 56.0 & 60.0 & 20.3 & 49.3 & 8.3 & 65.2 & 32.3 & 50.8 & 40.0 & 58.7 & 61.4 & 53.1 & 38.3 \\ 
        Yi-Coder-Chat & 9B & 50.6 & 45.7 & 54.7 & 28.3 & 47.0 & 40.0 & 42.2 & \underline{22.4} & 65.3 & 76.7 & 20.3 & 58.2 & 3.4 & 52.2 & 58.5 & 65.6 & 45.0 & 68.3 & 68.6 & 71.9 & 68.3 \\ 

        \midrule
        \multicolumn{23}{c}{\textbf{14B+ Models}} \\
        \midrule
        
        Qwen2.5-Instruct & 14B & 57.7 & 58.6 & 62.7 & 61.7 & 66.7 & 60.0 & 21.7 & 13.8 & 62.7 & 78.3 & 28.8 & 59.7 & 10.0 & 69.6 & 66.2 & 80.3 & 68.3 & 74.6 & 77.1 & 76.6 & 56.7 \\ 
        DS-Coder-V2-Lite-Instruct  & 2.4/16B & 56.7 & 10.0 & 56.2 & 43.3 & 56.1 & 81.5 & 50.6 & 10.3 & 58.7 & 76.7 & 28.8 & 68.7 & 17.2 & 65.2 & 63.1 & 72.1 & 71.7 & 76.2 & 80.0 & 60.9 & 81.7 \\ 
        Starcoder2-Instruct-v0.1 & 15B & 34.2 & 10.0 & 34.3 & 20.0 & 33.3 & 29.2 & 25.3 & 5.2 & 50.7 & 46.7 & 16.9 & 50.7 & 0.0 & 37.7 & 56.9 & 44.3 & 35.0 & 57.1 & 58.6 & 39.1 & 25.0 \\ 
        
        \midrule
        \multicolumn{23}{c}{\textbf{20B+ Models}} \\
        \midrule
        
        Codestral-v0.1 & 22B & 56.1 & 72.9 & 64.2 & 43.3 & 63.8 & 63.1 & 31.3 & 10.3 & 64.0 & 85.0 & 27.1 & 79.1 & 11.7 & 63.8 & 47.7 & 68.9 & 55.0 & 79.4 & 72.9 & 68.8 & 41.7 \\ 
        Qwen2.5-Instruct & 32B & 65.8 & 64.3 & 53.7 & 75.0 & 50.7 & 87.7 & 53.0 & 10.3 & 65.3 & \underline{93.3} & 32.2 & 74.6 & 13.3 & 81.2 & 75.4 & 90.2 & 80.0 & 85.7 & 82.9 & 84.4 & 58.3 \\ 
        Qwen2.5-Coder-Instruct & 32B & 68.2 & 78.6 & 60.9 & 75.0 & 56.1 & 83.1 & 44.6 & 13.8 & 61.3 & 91.7 & 33.9 & 85.1 & 22.4 & 82.6 & 64.6 & 91.8 & 80.0 & 79.4 & 82.9 & 82.8 & \underline{91.7} \\ 
        DS-Coder-Instruct & 33B & 57.7 & 65.7 & 59.4 & 46.7 & 50.0 & 70.8 & 39.8 & 19.0 & 65.3 & 75.0 & 28.8 & 73.1 & 10.3 & 58.0 & 55.4 & 73.8 & 61.7 & 79.4 & 68.6 & 78.1 & 70.0 \\ 
        CodeLlama-Instruct & 34B & 28.6 & 70.0 & 23.9 & 18.3 & 26.1 & 15.4 & 18.1 & 10.3 & 40.0 & 18.3 & 25.4 & 46.3 & 3.3 & 24.6 & 24.6 & 49.2 & 11.7 & 60.3 & 38.6 & 14.1 & 25.0 \\
        Meta-Llama-3-Instruct & 70B & 50.1 & 27.1 & 29.9 & 61.7 & 34.8 & 73.8 & 4.8 & 10.3 & 56.0 & 75.0 & 27.1 & 76.1 & 13.3 & 75.4 & 73.8 & 70.5 & 60.0 & 73.0 & 60.0 & 64.1 & 43.3 \\ 
        Meta-Llama-3.1-Instruct & 70B & 56.6 & 48.6 & 49.3 & 55.0 & 44.9 & 75.4 & 8.4 & 17.2 & 61.3 & 71.7 & 35.6 & 83.6 & 16.7 & 79.7 & 67.7 & 75.4 & 63.3 & 76.2 & 77.1 & 81.2 & 46.7 \\ 
        DS-V2.5 & 21/236B & 65.1 & 14.3 & 60.9 & 70.0 & 62.1 & 78.5 & 51.8 & 12.1 & 61.3 & 80.0 & 40.7 & 83.6 & 23.3 & 82.6 & 69.2 & 83.6 & 80.0 & 81.0 & 87.1 & 92.2 & 86.7 \\ 
        DS-V3 & 37/671B & 64.9 & 42.9 & 59.7 & 66.7 & 55.1 & 83.1 & 42.2 & 8.6 & 70.7 & 81.7 & 40.7 & 88.1 & 13.3 & 81.2 & 76.9 & 90.0 & 75.0 & 88.9 & 82.9 & 92.2 & 55.0 \\ 
        Qwen2.5-Instruct & 72B & 63.6 & 62.9 & 53.7 & 68.3 & 56.5 & 81.5 & 34.9 & 10.3 & 62.7 & 81.7 & 37.3 & 67.2 & 21.7 & 82.6 & 69.2 & 90.2 & 76.7 & 87.3 & 82.9 & 82.8 & 61.7 \\ 
        \rowcolor{green!8}  \textbf{\ourmethod{} (Our Method)} & 7B & 47.5 & 21.4 & 57.8 & 36.7 & 62.1 & 60.0 & 31.3 & 17.2 & 56.0 & 33.3 & 25.4 & 67.2 & 12.1 & 62.3 & 41.5 & 60.7 & 43.3 & 61.9 & 77.1 & 45.3 & 70.0 \\
  \bottomrule
  \end{tabular}
  }
   \label{tab:task_repair1}
  \vspace{-10pt}
  \end{table*}

%% file: table/code_ident.tex
\begin{table*}[t]
  \centering
  \caption{Accuracy of different models for Bug Identification tasks on \benchmark{}. The underlined numbers are the best scores for each language. ``Avg$_{all}$'' represents the average accuracy of all code languages.}
  \resizebox{1.0\textwidth}{!}{
  \begin{tabular}{l|cccccccccccccccccccccc}
  \toprule
\textbf{Model} &  Size & \textbf{Avg$_{all}$}  & C & C\# & CLISP & CPP & F\# & Go &HTML  & JS & Java  & Json   & Julia & MD & PHP & Pascal & Python & R & Ruby & Rust & Scala & Swift  \\ \midrule
        \multicolumn{23}{c}{\textbf{Closed-Source Models}} \\
        \midrule
        o1-preview & \faLock{} & \underline{37.0} & 34.3 & \underline{37.3} & 18.3 & \underline{36.2} & \underline{38.5} & \underline{47.0} & 25.9 & \underline{52.0} & \underline{31.7} & 13.6 & \underline{40.3} & 28.3 & \underline{33.3} & 32.3 & 52.5 & \underline{41.7} & \underline{38.1} & \underline{60.0} & \underline{35.9} & \underline{31.7}\\ 
        o1-mini & \faLock{} & 32.8 & 32.9 & 29.9 & \underline{25.0} & 30.4 & 23.1 & 38.6 & \underline{27.6} & 53.3 & 30.0 & 13.6 & 28.4 & 23.3 & 29.0 & 29.2 & 49.2 & 35.0 & 34.9 & 55.7 & 29.7 & 28.3\\ 
        GPT-4o-240806 & \faLock{} & 24.2 & 30.0 & 25.0 & 15.0 & 25.8 & 12.3 & 39.8 & 24.1 & 44.0 & 20.0 & 8.5 & 14.9 & 23.3 & 21.7 & 13.8 & 45.9 & 21.7 & 30.2 & 31.4 & 14.1 & 13.3 \\ 
        GPT-4o-mini-240718 & \faLock{} & 20.9 & 21.4 & 18.8 & 11.7 & 21.2 & 16.9 & 25.3 & 19.0 & 29.3 & 23.3 & 10.2 & 11.9 & 26.7 & 24.6 & 12.3 & 29.5 & 38.3 & 22.2 & 27.1 & 9.4 & 16.7\\ 
        Claude-3.5-sonnet-240620 & \faLock{} & 31.7 & \underline{44.3} & 26.6 & 20.0 & 24.2 & 26.2 & 44.6 & 19.0 & 45.3 & 23.3 & 13.6 & 35.8 & 25.0 & \underline{33.3} & 33.8 & 45.9 & 38.3 & 34.9 & 30.0 & 29.7 & 30.0\\ 
        Claude-3.5-sonnet-241022 & \faLock{} & 33.1 & 37.1 & 28.1 & 16.7 & 30.3 & 29.2 & 37.3 & 22.4 & 45.3 & 23.3 & 8.5 & 38.8 & 25.0 & \underline{33.3} & \underline{43.1} & \underline{55.7} & 40.0 & 36.5 & 38.6 & 34.4 & 30.0\\
        
        \midrule
        \multicolumn{23}{c}{\textbf{1B+ Models}} \\
        \midrule
        
        Qwen2.5-Instruct & 1.5B & 2.0 & 1.4 & 4.5 & 1.7 & 4.3 & 1.5 & 0.0 & 5.2 & 1.3 & 1.7 & 0.0 & 1.5 & 5.0 & 1.4 & 0.0 & 4.9 & 0.0 & 1.6 & 4.3 & 0.0 & 0.0 \\ 
        OpenCoder-Instruct & 1.5B & 4.2 & 0.0 & 0.0 & 18.3 & 1.4 & 0.0 & 2.4 & 1.7 & 1.3 & 0.0 & \underline{25.4} & 1.5 & 10.0 & 7.2 & 1.5 & 0.0 & 0.0 & 1.6 & 12.9 & 1.6 & 0.0 \\ 
        Qwen2.5-Instruct & 3B & 10.2 & 10.0 & 10.4 & 3.3 & 5.8 & 16.9 & 14.5 & 5.2 & 10.7 & 8.3 & 3.4 & 6.0 & 18.3 & 8.7 & 4.6 & 11.5 & 15.0 & 6.3 & 10.0 & 14.1 & 20.0\\
 
        \midrule
        \multicolumn{23}{c}{\textbf{7B+ Models}} \\
        \midrule
        
        Qwen2.5-Coder-Instruct & 7B & 8.4 & 11.4 & 4.7 & 8.3 & 3.0 & 6.2 & 15.7 & 8.6 & 14.7 & 8.3 & 10.2 & 7.5 & 13.8 & 4.3 & 1.5 & 16.4 & 8.3 & 11.1 & 8.6 & 0.0 & 3.3\\ 
        Meta-Llama-3-Instruct & 8B & 3.0 & 2.9 & 4.5 & 3.3 & 2.9 & 0.0 & 3.6 & 0.0 & 8.0 & 6.7 & 1.7 & 0.0 & 0.0 & 4.3 & 0.0 & 9.8 & 1.7 & 0.0 & 7.1 & 0.0 & 1.7 \\ 
        Meta-Llama-3.1-Instruct & 8B & 5.4 & 7.1 & 10.4 & 3.3 & 11.6 & 0.0 & 7.2 & 1.7 & 5.3 & 3.3 & 1.7 & 6.0 & 3.3 & 8.7 & 3.1 & 9.8 & 1.7 & 6.3 & 4.3 & 1.6 & 10.0\\ 
        Yi-Coder-Chat & 9B & 8.7 & 25.7 & 9.4 & 5.0 & 9.1 & 4.6 & 10.8 & 5.2 & 12.0 & 11.7 & 1.7 & 16.4 & 6.9 & 5.8 & 4.6 & 14.8 & 5.0 & 3.2 & 5.7 & 7.8 & 5.0\\

        \midrule
        \multicolumn{23}{c}{\textbf{20B+ Models}} \\
        \midrule

        Codestral-v0.1 & 22B & 16.2 & 21.4 & 19.4 & 10.0 & 21.7 & 6.2 & 31.3 & 12.1 & 20.0 & 18.3 & 6.8 & 16.4 & 20.0 & 7.2 & 10.8 & 32.8 & 8.3 & 20.6 & 14.3 & 12.5 & 6.7\\ 
        Qwen2.5-Instruct & 32B & 19.4 & 28.6 & 25.4 & 10.0 & 23.2 & 9.2 & 34.9 & 12.1 & 24 & 18.3 & 10.2 & 26.9 & \underline{30.0} & 11.6 & 10.8 & 32.8 & 13.3 & 25.4 & 14.3 & 9.4 & 10.0 \\ 
        Qwen2.5-Coder-Instruct & 32B & 23.6 & 30.0 & 25.0 & 13.3 & 31.8 & 15.4 & 37.3 & 22.4 & 28.0 & 16.7 & 11.9 & 31.3 & 29.3 & 18.8 & 21.5 & 36.1 & 36.7 & 28.6 & 20.0 & 4.7 & 6.7\\ 
        DS-Coder-Instruct & 33B & 12.3 & 14.3 & 15.6 & 0.0 & 15.2 & 6.2 & 15.7 & 12.1 & 22.7 & 8.3 & 6.8 & 11.9 & 27.6 & 10.1 & 10.8 & 24.6 & 6.7 & 17.5 & 11.4 & 4.7 & 1.7 \\ 
        Qwen2.5-Instruct & 72B & 17.6 & 28.6 & 16.4 & 13.3 & 18.8 & 7.7 & 25.3 & 19.0 & 26.7 & 15.0 & 8.5 & 20.9 & 28.3 & 13.0 & 6.2 & 26.2 & 21.7 & 22.2 & 12.9 & 7.8 & 10.0\\ 
        DS-Coder-V2.5 & 21/236B & 19.2 & 21.4 & 17.2 & 11.7 & 16.7 & 13.8 & 32.5 & 19.0 & 29.3 & 18.3 & 5.1 & 13.4 & 20.0 & 8.7 & 15.4 & 37.7 & 15.0 & 23.8 & 25.7 & 17.2 & 15.0\\ 
        \rowcolor{green!8}  \textbf{\ourmethod{} (Our Method)} & 7B & 2.3 & 2.9 & 4.7 & 3.3 & 4.5 & 0.0 & 1.2 & 6.9 & 8.0 & 0.0 & 0.0 & 0.0 & 0.0 & 1.4 & 3.1 & 4.9 & 0.0 & 0.0 & 2.9 & 0.0 & 1.7\\
          
  \bottomrule
  \end{tabular}}
  \label{tab:ident}
  \vspace{-5pt}
  \end{table*}

%% file: table/code_loc.tex
\begin{table*}[t]
  \centering
  \caption{Accuracy of different models for Bug Localization tasks on \benchmark{}. The underlined numbers are the best scores for each language. ``Avg$_{all}$'' represents the average scores of all code languages.}
  \resizebox{1.0\textwidth}{!}{
  \begin{tabular}{l|cccccccccccccccccccccc}
  \toprule

\textbf{Model} &  Size & \textbf{Avg$_{all}$}  & C & C\# & CLISP & CPP & F\# & Go &HTML  & JS & Java  & Json   & Julia & MD & PHP & Pascal & Python & R & Ruby & Rust & Scala & Swift  \\ \midrule
        \multicolumn{23}{c}{\textbf{Closed-Source Models}} \\
        \midrule
        o1-preview & \faLock{} & 64.9  & 72.9  & 50.7  & 30.0  & 62.3  & 60.0  & 49.4  & 74.1  & 72.0  & 66.7 & \underline{79.7}  & 74.6  & 50.0 & 79.7 & 55.4 & \underline{86.9} & 71.7 & 66.7 & 54.3 & 71.9 & \underline{73.3} \\ 
        o1-mini & \faLock{} & \underline{68.1}  & \underline{78.6}  & \underline{65.7}  & 41.7  & 63.8  & \underline{67.7}  & 47.0  & 69.0  & \underline{81.3}  & \underline{71.7} & 67.8  & \underline{76.1}  & 53.3 & \underline{84.1} & 60.0 & 78.7 & 78.3 & \underline{79.4} & 60.0 & \underline{73.4} & 66.7 \\ 
        GPT-4o-240806 & \faLock{} & 56.1  & 60.0  & 53.1  & 30.0  & 54.5  & 61.5  & 47.0  & 65.5  & 65.3  & 60.0  & 62.7  & 50.7  & 46.7  & 58.0  & 53.8  & 63.9  & 61.7  & 66.7  & 55.7  & 57.8  & 48.3\\ 
        GPT-4o-mini-240718 & \faLock{} & 36.8  & 51.4  & 28.1  & 23.3  & 25.8  & 32.3  & 37.3  & 58.6  & 54.7  & 36.7 & 45.8  & 20.9  & 40.0  & 30.4  & 26.2  & 44.3  & 36.7  & 50.8  & 37.1  & 23.4  & 31.7 \\ 
        Claude-3.5-sonnet-240620 & \faLock{} & 62.9  & 74.3  & 62.5  & \underline{61.7}  & 60.6  & 50.8  & \underline{59.0}  & 67.2  & 73.3  & 63.3 & 69.5  & 71.6  & 55.0  & 60.9  & 58.5  & 68.9  & 68.3  & 58.7  & 58.6  & 57.8  & 56.7  \\ 
        Claude-3.5-sonnet-241022 & \faLock{} & 64.2  & 67.1  & 60.9  & 58.3  & 59.1  & 55.4  & \underline{59.0}  & 69.0  & 73.3  & 56.7 & 72.9  & 73.1  & \underline{65.0}  & 60.9  & \underline{66.2}  & 68.9  & 76.7  & 55.6  & \underline{67.1}  & 57.8  & 61.7 \\ 
        
        \midrule
        \multicolumn{23}{c}{\textbf{1B+ Models}} \\
        \midrule
        
        Qwen2.5-Instruct & 1.5B & 22.8  & 22.9  & 40.3  & 13.3  & 21.7  & 9.2  & 26.5  & 8.6  & 26.7  & 36.7& 16.9  & 34.3  & 21.7 & 15.9 & 21.5 & 19.7 & 30.0 & 12.7 & 24.3 & 15.6 & 33.3\\ 
        OpenCoder-Instruct & 1.5B & 10.5  & 1.4  & 17.9  & 10.0  & 5.8  & 6.2  & 16.9  & 5.2  & 25.3  & 13.3 & 8.5  & 10.4  & 8.3 & 14.5 & 15.4 & 6.6 & 1.7 & 6.3 & 14.3 & 7.8 & 8.3 \\ 
        Qwen2.5-Instruct & 3B & 21.4  & 30.0  & 14.9  & 16.7  & 13.0  & 23.1  & 18.1  & 29.3  & 21.3  & 31.7  & 27.1  & 19.4  & 21.7 & 20.3 & 20.0 & 36.1 & 26.7 & 17.5 & 12.9 & 23.4 & 8.3\\ 
 
        \midrule
        \multicolumn{23}{c}{\textbf{7B+ Models}} \\
        \midrule
        
        Qwen2.5-Coder-Instruct & 7B & 26.8  & 42.9  & 21.9  & 16.7  & 27.3  & 23.1  & 31.3  & 8.6  & 32.0  & 33.3 & 32.2  & 25.4  & 10.3  & 23.2  & 32.3  & 41.0  & 18.3  & 36.5  & 28.6  & 31.2  & 13.3 \\ 
        Meta-Llama-3-Instruct & 8B & 7.8  & 8.6  & 6.0  & 3.3  & 5.8  & 1.5  & 10.8  & 12.1  & 10.7  & 11.7 & 18.6  & 6.0  & 18.3 & 4.3 & 3.1 & 3.3 & 1.7 & 6.3 & 11.4 & 10.9 & 1.7 \\ 
        Meta-Llama-3.1-Instruct & 8B & 7.0  & 7.1  & 9.0  & 10.0  & 5.8  & 3.1  & 12.0  & 17.2  & 0.0  & 6.7 & 27.1  & 4.5  & 13.3 & 2.9 & 1.5 & 3.3 & 3.3 & 0.0 & 8.6 & 7.8 & 0.0  \\ 
        Yi-Coder-Chat & 9B & 29.5  & 42.9  & 40.6  & 20.0  & 34.8  & 29.2  & 22.9  & 0.0  & 60.0  & 40.0 & 18.6  & 28.4  & 1.7  & 42.0  & 30.8  & 6.6  & 26.7  & 36.5  & 35.7  & 25.0  & 33.3 \\ 

        \midrule
        \multicolumn{23}{c}{\textbf{20B+ Models}} \\
        \midrule
        
        Codestral-v0.1 & 22B & 43.6  & 52.9  & 41.8  & 35.0  & 44.9  & 43.1  & 42.2  & 32.8  & 62.7  & 48.3 & 32.2  & 53.7  & 20.0 & 50.7 & 43.1 & 60.7 & 40.0 & 47.6 & 38.6 & 42.2 & 31.7  \\ 
        Qwen2.5-Instruct & 32B & 58.4  & 68.6  & 50.7  & 33.3  & \underline{65.2}  & 55.4  & 49.4  & \underline{75.9}  & 65.3  & 60.0 & 66.1  & 68.7  & 53.3 & 56.5 & 50.8 & 62.3 & 68.3 & 61.9 & 58.6 & 51.6 & 46.7 \\ 
        Qwen2.5-Coder-Instruct & 32B & 59.4  & 81.4  & 50.0  & 46.7  & \underline{65.2}  & 64.6  & 63.9  & 13.8  & 72.0  & 68.3 & 52.5  & 68.7  & 12.1  & 66.7  & 55.4  & 73.8  & \underline{85.0}  & 63.5  & 60.0  & 59.4  & 50.0 \\ 
        DS-Coder-Instruct & 33B & 17.8  & 28.6  & 12.5  & 8.3  & 19.7  & 16.9  & 20.5  & 1.7  & 38.7  & 33.3 & 8.5  & 20.9  & 3.4  & 13.0  & 12.3  & 1.6  & 10.0  & 27.0  & 31.4  & 15.6  & 21.7 \\ 
        Qwen2.5-Instruct & 72B & 57.4  & 74.3  & 44.8  & 43.3  & 44.9  & 66.2  & 55.4  & 65.5  & 64.0  & 68.3 & 61.0  & 67.2  & 45.0 & 50.7 & 47.7 & 68.9 & 61.7 & 60.3 & 50.0 & 59.4 & 50.0 \\ 
        DS-Coder-V2.5 & 21/236B & 52.3  & 75.7  & 50.0  & 28.3  & 56.1  & 40.0  & 54.2  & 63.8  & 64.0  & 68.3 & 59.3  & 44.8  & 43.3  & 58.0  & 27.7  & 70.5  & 53.3  & 34.9  & 47.1  & 57.8  & 46.7 \\ 
        \rowcolor{green!8} \textbf{\ourmethod{} (Our Method)} & 7B & 18.7  & 30.0  & 21.9  & 3.3  & 15.2  & 10.8  & 20.5  & 8.6  & 30.7  & 36.7 & 28.8  & 23.9  & 8.6  & 18.8  & 15.4  & 13.1  & 5.0  & 33.3  & 20.0  & 15.6  & 6.7\\

  \bottomrule
  \end{tabular}}
  \label{tab:task_loc}
  \vspace{-10pt}
  \end{table*}